\documentclass{article}

\PassOptionsToPackage{numbers, compress, sort}{natbib}

\usepackage{natbib}

\usepackage[preprint]{neurips_2020}

\usepackage[utf8]{inputenc} % allow utf-8 input
\usepackage{wrapfig,lipsum,booktabs}

\usepackage[T1]{fontenc}    % use 8-bit T1 fonts
\usepackage{hyperref}       % hyperlinks
\usepackage{url}            % simple URL typesetting
\usepackage{booktabs}       % professional-quality tables
\usepackage{amsfonts}       % blackboard math symbols
\usepackage{nicefrac}       % compact symbols for 1/2, etc.
\usepackage{microtype}      % microtypography
\usepackage{siunitx}        % SI units
\usepackage{dsfont}         % double-stroke font for maths
\usepackage{amsmath}        % mathematical typesetting
\usepackage{tikz}           % graphics
\usepackage{changes}
\usepackage{lipsum}
\usepackage[font={footnotesize}]{caption}
\usepackage{bold-extra}
\usepackage{xcolor}
\usepackage{multirow}
\usepackage{csvsimple}
\usepackage{titlesec}
\usepackage{pbox}
\usepackage[scaled=0.85]{beramono}
\usepackage{pdfpages}
\usepackage{placeins}
\usepackage{comment}
\usepackage{graphicx}
\usepackage{enumitem}
\usepackage{appendix}
\usepackage{subcaption}
\usepackage{wrapfig,booktabs}
\makeatletter

\usepackage{hyperref}
\definecolor{myblue}{rgb}{0, 0, 0.7}
\hypersetup{colorlinks=true,
            linkcolor = myblue,
            urlcolor  = myblue,
            citecolor = myblue}

\newcommand{\reals}{\mathds{R}}

\DeclareMathOperator{\MMD}{MMD}

\title{Accelerating COVID-19 Differential Diagnosis with Explainable Ultrasound Image Analysis}

\author{
 Jannis Born$^{1,*}$
  \And
  Nina Wiedemann$^{2,*}$
\AND
  Gabriel Br\"andle$^{3}$
\And
  Charlotte Buhre$^{4}$
\And
  Bastian Rieck$^{1}$ 
\And
  Karsten Borgwardt$^{1}$
  \AND
  \small{$^{*}$\textmd{Shared first-authorship. \texttt{\{jborn,wnina\}@ethz.ch}}}\\
  \small\textsc{$^{1}$Department of Biosystems Science and Engineering, ETH Zurich, Switzerland}\\
    \small\textsc{$^{2}$Department of Computer Science, ETH Zurich, Switzerland}\\
    \small\textsc{$^{3}$Hisrslanden Clinique des Grangettes, Geneva, Switzerland}\\
    \small\textsc{$^{4}$Medizinische Hochschule Brandenburg Theodor Fontane, Germany}\\
    }
\begin{document}

\maketitle

\begin{abstract}
  % Medical imaging techniques such as X-rays or CT scans were
  % demonstrated promising performances in some initial pilot studies.
  %
  %With the rapid development of COVID-19 into a global pandemic, there is
  %an ever-increasing need for cheap, fast, and reliable tools that can
  %assist clinicians in their diagnosis efforts.
  %
  Controlling the COVID-19 pandemic largely hinges upon the existence of fast, safe, and highly-available diagnostic tools.
  Ultrasound, in contrast to CT or X-Ray, has many practical advantages and can serve as a globally-applicable first-line examination technique. % for suspected infections.
  %Fast, safe, and highly-available diagnostic tools are key to control the COVID-19 pandemic. 
  %As an established method with significant practical advantages, lung ultrasound may serve as a globally-applicable first-line examination method.
  %In contrast to CT or X-Ray, ultrasound has many practical advantages and may serve as a globally-applicable first-line examination method for COVID-19 suspects.
  % Ultrasound as an established method with significant practical advantages may -- in contrast to CT or X-Ray -- 
  % may be a key part of a roadmap for globally applicable first-line examination.
  % serve as a globally applicable first-line examination method. 
  % Given the difficulties of pattern recognition for clinicians, computer vision can increase the reliability of this potentially powerful testing method.
  % Considering significant practical advantages of ultrasound, but difficulties for doctors in pattern recognition, computer vision can leverage the applicability of a potentially powerful testing method and, in contrast to CT or X-Ray, may provide a roadmap for a globally applicable first-line examination method.
  % 
  % We train a convolutional neural network to classify a novel ultrasound dataset of 107 videos with three classes, COVID-19, bacterial pneumonia, and healthy controls. The data was curated and approved by medical experts and is released as the largest publicly available lung ultrasound dataset. On this data, our model 
  %
  We provide the largest publicly available lung ultrasound (US) dataset for COVID-19 consisting of 106 videos from three classes~(COVID-19, bacterial pneumonia, and healthy controls); curated and approved by medical experts.
  On this dataset, we perform an in-depth study of the value of deep learning methods for differential diagnosis of COVID-19.
  We propose a frame-based convolutional neural network that %achieves an accuracy of $0.9\pm0.02$ and
  correctly classifies COVID-19 US videos with a sensitivity of $0.98\pm0.04$ and a specificity of $0.91\pm0.08$ (frame-based sensitivity $0.93\pm0.05$, specificity $0.87\pm0.07$). 
  We further employ class activation maps for the spatio-temporal localization of pulmonary biomarkers, which we subsequently
  validate for human-in-the-loop scenarios in a blindfolded 
  %human-in-the-loop 
  study with medical experts.
  %
  %Class activation maps are demonstrated as a potent tool for spatio-temporal localization of pulmonary biomarkers and are validated in a blind-folded human-in-the-loop study by medical experts.
  Aiming for scalability and robustness, we perform ablation studies comparing mobile-friendly, frame- and video-based architectures and show reliability of the best model by aleatoric and epistemic uncertainty estimates.
  %
  % Last, we validate our model on an independent test dataset of 66 videos and report promising performance.
  We hope to pave the road for a community effort toward an 
  % to contribute toward an
  accessible, efficient and interpretable screening method and we have started to work on a clinical validation of the proposed method.
    %
    % BR: Maybe put this into the conclusion?
    % Overall, our paper demonstrates the potential of employing efficient deep learning techniques to ease the testing and detection burden of
    % \mbox{COVID-19}.
  %
  %Our paper together with  a proof-of-concept web-service thus demonstrates the potential of employing efficient deep learning techniques to ease the testing and detection burden of
  %\mbox{COVID-19}.
  Data and code 
  % of our open-access initiative
  are publicly available.

  % Medical imaging technologies such as computed tomography~(CT) or X-ray can take a key role in complementing conventional diagnostic tools from molecular biology, and, using deep learning techniques, several  automated systems demonstrated promising performances.
  %
  % In this paper, we advocate for a prominent role of point-of-care ultrasound imaging to guide COVID-19 detection. Ultrasound does not require exposure to radiation, is non-invasive, and ubiquitously found in  medical facilities around the globe.
  %
  % In this paper, we assemble a novel ultrasound dataset based on different online sources. We then show how to train a convolutional neural network that is capable of predicting \mbox{COVID-19} with a sensitivity of $0.96$ and a specificity of $0.79$. In addition to probing the model to analyse its interpretability, we also describe a web service for live testing of the model.
  %
  % Our paper thus demonstrates the potential of employing efficient deep learning techniques to ease the testing and detection burden of \mbox{COVID-19}.
\end{abstract}

\begin{figure}[!htb]
\includegraphics[width=\linewidth]{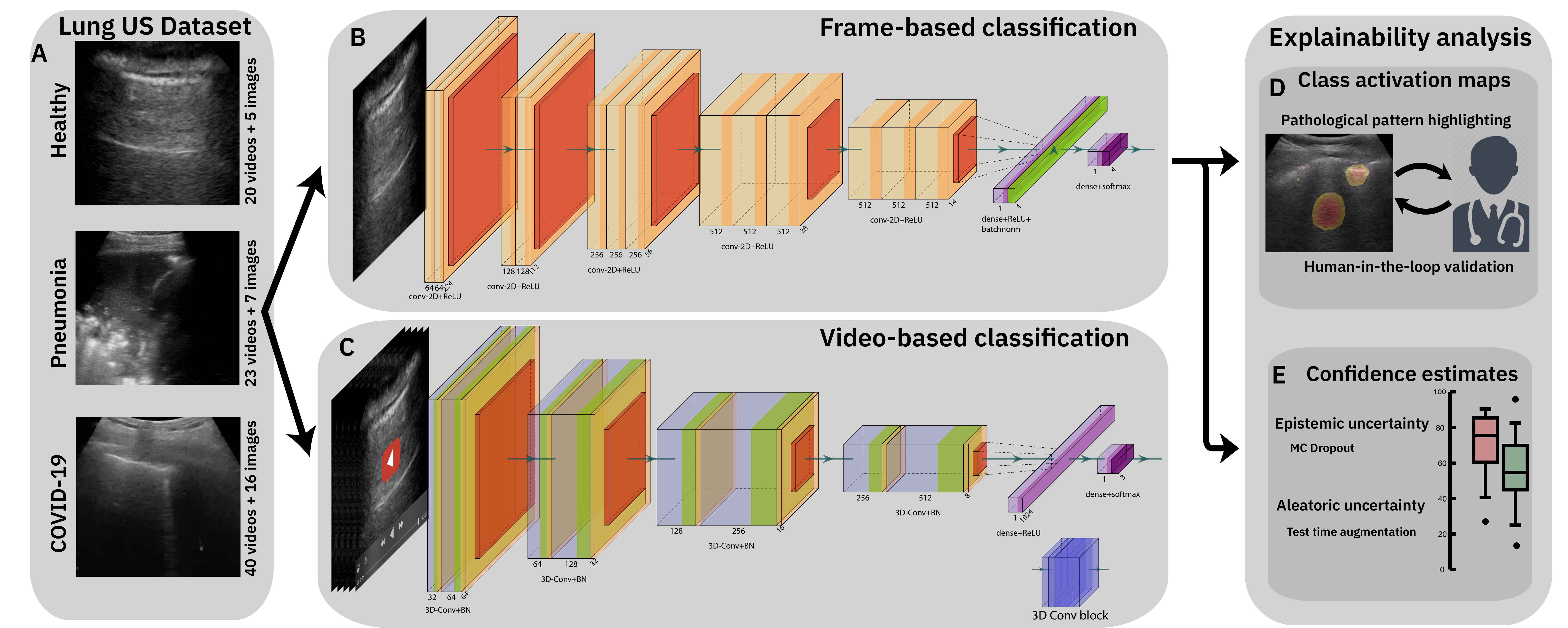}
\caption{
\textbf{Flowchart of our contribution.}
\textbf{A:} 3 samples from our public COVID-19 lung US dataset.
\textit{Top}: Healthy lung with horizontal A-lines,~\textit{Middle}: pneumonia infected lung with alveolar consolidations,~\textit{Bottom}: SARS-CoV-2 infected lung with subpleural consolidation and a focal B-line.
\textbf{B},\textbf{C}: We fine-tune and compare frame- and video-based CNNs on this new dataset and demonstrate the feasibility of differential diagnosis from ultrasound.
\textbf{D}: Class activation maps highlight patterns that drove the model's decision and are reviewed and evaluated for diagnostic value by medical experts.
\textbf{E}: Uncertainty techniques are employed and shown to equip the model with the ability to recognize samples with high error probability.
}
\label{fig:overview}
\end{figure}

\section{Introduction}

To date, SARS-CoV-2 has infected several millions and COVID-19 has killed hundreds of thousands around the globe.
Its long incubation time calls for fast, accurate, and reliable techniques for early disease diagnosis to successfully fight the spread~\cite{li2020early}.
The standard genetic test~(RT-PCR), suffers from a processing time of up to 2 days~\citep{mei2020artificial}, several publications reported sensitivity as low as 70\%~\cite{kanne2020essentials,ai2020correlation} and a recent meta-analysis estimated the \emph{false negative} rate to be at least 20\% over the course of the infection~\citep{kucirka2020variation}.
Medical imaging has great potential to complement the diagnostic process as a fast assessment tool that guides further PCR-testing, especially in triage situations~\citep{dong2020role}.
Currently, CT scans are the gold standard for pneumonia~\cite{bourcier2014performance} and are considered relatively reliable for COVID-19 diagnosis~\cite{bao2020covid,fang2020sensitivity,ai2020correlation}, although a significant amount of patients exhibit normal CT scans~\citep{yang2020patients}.
However, performing CT is expensive and highly irradiating, posing risks of infection for patients and staff, requires extensive sterilization~\cite{mossa2020radiology}, and is reserved for developed countries; there are only $\sim$30k CT scanner globally~\citep{castillo2012industry}.
Chest X-ray~(CXR) scans are still the first line examination, despite some reports of low specificity and sensitivity for COVID-19~(for example, \citet{weinstock2020chest} found 89\% normal CXR in 493 COVID-19 patients).
Ultrasound~(US), by contrast, is a cheap, safe, non-invasive and repeatable technique that can be performed with portable devices at patient bedside and is ubiquitously available around the globe. 
Over the last two decades, ultrasound became an established tool to diagnose pulmonary diseases~\citep{lichtenstein2004comparative,chavez2014lung,pagano2015lung}, has been forecast to replace radiographic techniques~\citep{bourcier2016lung}, was demonstrated to be superior to CXR for pulmonary diseases~\citep{reali2014can,claes2017performance}, and started to replace X-Ray as first-line examination~\citep{abdalla2016lung,brogi2017could}.

In the COVID-19 outbreak, a growing body of evidence for disease-specific patterns in US (e.g. B-lines and subpleural consolidations) has led to advocacy for an amplified role of US from the research community~\citep{buonsenso2020covid,smith2020point,lepri2020emerging,sultan2020review} and radiologists reported great agreement between US and CT findings for COVID-19 infections~\cite{peng2020findings,fiala2020ultrasound}.
Moreover, in triage situations or in third-world countries, where CT, PCR and CXR are not available, US was demonstrated to be a valuable patient stratification technique for pneumonia~\cite{ellington2017lung,amatya2018diagnostic}.
This gives US, in contrast to other imaging techniques, the potential to become a globally applicable first-line examination method~\citep{stewart2020trends}.
However, the relevant pattern are hard to discern for humans~\citep{ng2020imaging,tutino2010time}, calling into play medical image analysis based on machine learning technique as a decision support tool for physicians.
Here, we provide the first study of automatic lung ultrasound analysis for differential diagnosis of bacterial and viral pneumonia; aiming to develop a medical decision support tool.

\paragraph{Related work.}
Literature on exploiting medical image analysis and computer vision techniques to classify or segment CT or CXR data of COVID-19 patients recently exploded~(for reviews, see \citet{ulhaq2020computer,shi2020review}, for a list of public data sources see~\citet{kalkreuth2020covid}).
For example, in an early study, \citet{butt2020deep} reported a sensitivity of 98\% (specificity~92\%) in a binary classification on CT scans from 110 COVID-19 patients, while \citet{mei2020artificial} very recently achieved equal sensitivity~(but lower specificity) compared to senior radiologists in detecting COVID-19 from CT and clinical information of 279 patients. 
US instead has been neglected heavily by the ML community~\citep{born2020role}; only the Italian COVID-19 Lung Ultrasound~(\texttt{ICLUS}) project has proposed a deep learning approach for a severity assessment of COVID-19 from ultrasound data~\citep{roy2020deep}.
The work convincingly predicts disease severity and segments COVID-19 specific patterns, building up on their previous work on localizing B-lines~\cite{van2019localizing}.
The paper claims to release a dataset of annotated COVID-19 cases, but to date, no annotations are available.
While this effort is highly relevant for disease monitoring, it is not directly applicable for first-line diagnosis, where the main problem lies in \emph{distinguishing} COVID-19 from other pneumonia.
We aim to close this gap with our approach to classify COVID-19, healthy, and pneumonia point-of-care ultrasound (POCUS) images.

\paragraph{Our contributions.}
\autoref{fig:overview} depicts a graphical overview of our contributions.
We provide the largest publicly-available dataset of lung US recordings consisting of 106 videos.
This dataset is heterogeneous and mostly from public sources, but was curated manually and approved by a medical doctor.
We further take a first step towards a tool for differential diagnosis of pulmonary diseases, here especially focused on bacterial and viral pneumonia such as COVID-19.
An earlier version of our dataset alongside some preliminary results, we already presented in~\citep{born2020pocovid}.
Without deprivation of novelty, we here
% Specifically, we
% add to the existing literature of CT and CXR image analysis for COVID-19 by
demonstrate that competitive performance can be achieved from raw US recordings, thereby challenging the current focus on irradiating imagining techniques. 
Moreover, we employ explainability techniques such as class activation maps or uncertainty estimates and present a roadmap towards an automatic detection system that can \emph{segment} and \emph{highlight} relevant spatio-temporal patterns. 
Such a system could not only lead to superior diagnostic performance, as was partially shown for CT~\citep{mei2020artificial}, but can also reduce the time doctors require to make a diagnosis~\cite{shan2020lung}.
Our approach is of evident need because physicians must be trained thoroughly to reliably differentiate COVID-19 from pneumonia~\cite{ng2020imaging}, making it necessary to use powerful deep learning to develop a system that can complement the work of physicians in a timely manner.

\section{A lung ultrasound dataset for  COVID-19 detection}
We provide the to-date largest pre-processed and publicly available lung POCUS dataset\footnote{\url{https://github.com/jannisborn/covid19_pocus_ultrasound/tree/master/data}}, comprising samples of COVID-19 patients, pneumonia-infected lungs and healthy patients.
As shown in~\autoref{tab:data_numbers}, we collected and gathered $139$ recordings ($106$ videos + $33$ images) recorded with either convex or linear probes, where the latter is a higher frequency probe yielding more superficial images.
\begin{wraptable}{r}{5.5cm}
\scalebox{0.4}{
\resizebox{\textwidth}{!}{%
\begin{tabular}{cccccl}
\cline{2-6}
\textbf{}                                          & \multicolumn{2}{c}{\textbf{Convex}}                                        & \multicolumn{2}{c}{\textbf{Linear}} &              \\ \cline{2-6} 
\multicolumn{1}{l}{}                              & \multicolumn{1}{l}{\textbf{Vid.}} & \multicolumn{1}{l}{\textbf{Img.}} & \textbf{Vid.}   & \textbf{Img.}  &  \textbf{Sum}            \\ \hline
\multicolumn{1}{c}{\textbf{COVID}}               & 40                                   & 16                                   & 4                 & 3                & \textbf{63}  \\ \hline
\multicolumn{1}{c}{\textbf{BP}} & 23                                   & 7                                    & 2                 & 2                & \textbf{34}  \\ \hline
\multicolumn{1}{c}{\textbf{VP}}     & 3                                    &  --                                    & 4                 & --                  & \textbf{7}   \\ \hline
\multicolumn{1}{c}{\textbf{Healthy}}             & 20                                   & 5                                    & 10                & --                 & \textbf{35}  \\ \hline
\multicolumn{1}{c}{\textbf{Sum}}                 & \textbf{86}                          & \textbf{28}                          & \textbf{20}       & \textbf{5}       & \textbf{139} \\ \hline
\end{tabular}%
}
}
\caption{Number of videos and images in our dataset, per class and probe. BP is bacterial pneumonia, VP is viral pneumonia.}
\vspace{-3mm}
\label{tab:data_numbers}
\end{wraptable} 
Our sources comprise community platforms, open medical repositories, health-tech companies, other scientific literature, and data recorded by healthy volunteers from our team.
Main sources of data were
\href{https://www.grepmed.com}{grepmed.com},
\href{https://thepocusatlas.com}{thepocusatlas.com},
\href{https://www.butterflynetwork.com}{butterflynetwork.com} and 
\href{https://radiopaedia.org}{radiopaedia.org}. %~ (for details, see appendix~\ref{appendix:data}).
All samples of our database were annotated and approved by a medical doctor; moreover notes on the visible patterns in each video~(e.g.\ B-Lines or consolidations) were added.
In all collected videos of COVID-19 and pneumonia, disease-specific patterns are visible.
The dataset is heterogeneous in terms of resolution, frame rates, the conducted lung US protocol, the devices used, and little to no meta data about the patients is available. For more details about the dataset and metadata, see the release on GitHub.

\section{Differential diagnosis of COVID-19 with lung ultrasound}

\subsection{Experimental setup}
\paragraph{Data processing.}
All experiments are conducted on data recorded with convex ultrasound probes, the standard probe for lung assessment that allows to see deeply into the lung~\citep{lichtenstein2015lung}.
%We use several neural network architectures to tackle the present classification task, and implement class activation mapping methods and uncertainty techniques for explainability.
%
%\paragraph{Data processing.}
We manually processed all convex ultrasound recordings and split them into images at a frame rate of 3Hz~(with maximal 30 frames per video), leading to a database of 693 COVID-19, 377 bacterial pneumonia, and 295 healthy control images. For examples see~\autoref{fig:overview}A.
All images were cropped to a quadratic window excluding measure bars and texts and artifacts on the borders before they were resized to $224 \times 224$ pixels.
%
%\paragraph{Data splitting.}
Apart from the independent test data, all reported results were obtained in a 5-fold stratified cross validation. 
It was ensured that the frames of a single video are present within a single fold only, and that the number of samples per class is similar in all folds.
% \footnote{As a consequence though, the sizes of each fold vary, for example with 112 COVID-19 images in the smallest and 197 in the largest split.}
All models were trained to classify images as COVID-19, pneumonia, healthy, or uninformative.
The latter consists of ImageNet pictures as well as neck ultraosund data; we added these picture for the purpose of detecting out-of-distribution data~(thus making the model more robust). This is particularly relevant for public web-based inference services. 
In this paper, we present all results \emph{omitting the uninformative class}, as it is not relevant for the analysis of differential diagnosis performance and would bias the results~(please refer to appendix~\ref{appendix:uniformative} for results including uninformative data). Furthermore, we use data augmentation techniques~(horizontal and vertical flips, rotations up to 10$^{\circ}$ and translations of up to 10\%)  to diversify the dataset and prevent overfitting.
\paragraph{Frame-based models.}
Our backbone neural architecture is a \texttt{VGG-16}~\cite{Simonyan15} that is compared to \texttt{NasNET Mobile}, a light-weight alternative ~\cite{zoph2018learning} that uses less than $\nicefrac{1}{3}$ of the parameters of \texttt{VGG} and was optimized for applications on portable devices.
Both models are pre-trained on \texttt{Imagenet} and fine-tuned on the frames sampled from the videos.
Specifically, we use two variants of \texttt{VGG-16} that we name \texttt{VGG} and \texttt{VGG-CAM}.
\texttt{VGG-CAM} has a single dense layer following the convolutions, thus enabling the usage of plain CAMs, class activation maps~\citep{zhou2016learning}, whereas \texttt{VGG} has an additional dense layer with \texttt{ReLU} activation and batch normalization.

Considering the recent work of \citet{roy2020deep} on lung US segmentation and severity prediction for COVID-19, we investigated whether a segmentation-targeted network can also add value to the prediction in differential diagnosis.
We implemented two approaches building upon the pre-trained model of \citet{roy2020deep}, an ensemble of three separate \texttt{U-Net}-based models~(\texttt{U-Net}, \texttt{U-Net++}, and \texttt{DeepLabv3+}, with a total of $\sim19.5$M parameters).
First, \texttt{VGG-Segment} is identical to \texttt{VGG}, however instead of training on the raw US data, we train on the segmented images from the ensemble (see example in Appendix~\ref{appendix:segment}).
Although it might seem unconventional, we hypothesized that the colouring entails additional information that might simplify classification.
Secondly, in~\texttt{Segment-Enc} the bottleneck layer of each of the three models is used as a feature encoding of the images, resulting in $560$ filter maps that are fed through two dense layers of size $512$ and $256$ respectively.
The encoding weights are fixed during training. Both settings are compared to the other models that directly utilize the raw images.
For more details on the architectures and the training procedure, please refer to appendix~\ref{appendix:arch}.

\paragraph{Video-based model.}
In comparison to a na\"ive, frame-based video classifier~(obtained by averaging scores of all frames), we also investigate \texttt{Models Genesis}, a generic model for 3D medical image analysis pretrained on lung CT scans~\citep{zhou2019models}.
For \texttt{Models Genesis}, the videos are split into chunks of $5$ frames each, sampled at a frame rate of 5Hz.
5-fold cross validation is performed using the same split as for frame-based classifiers.
Individual images were excluded, leaving aside 86 videos (from which 10 were excluded due to too many frames with artifacts such as moving pointers) which were split into 292 video chunks.

\subsection{Frame-based experiments}

% \input{compare_table}
% Table spreading both columns
\newcommand\Tstrut{\rule{0pt}{2.6ex}}        % = `top' strut
\newcommand\Bstrut{\rule[-0.9ex]{0pt}{0pt}}

\begin{table*}[t]
  \centering
\scalebox{0.9}{
\begin{tabular}{llccccc}
\toprule
{} &    \textbf{Class} & \textbf{Recall} &  \textbf{Precision} &   \textbf{F1-score} &     \textbf{Specificity} &   \textbf{MCC} 
\\ \midrule 
\multirow{3}{*}{\pbox{20cm}{\textbf{VGG}\\  \small{Acc.: 0.90, Bal.: \textbf{0.90}} \\ \# Par.: 14 747 971}
} & COVID-19 & $ 0.89 \pm {\scriptstyle 0.06 }$ & $ \textbf{0.91} \pm {\scriptstyle 0.05 }$ & $ 0.90 \pm {\scriptstyle 0.03 }$ & $ \textbf{0.92} \pm {\scriptstyle 0.04 }$ & $ 0.80 \pm {\scriptstyle 0.03 } $ \\
& Pneumonia & $ 0.94 \pm {\scriptstyle 0.06 }$ & $ 0.93 \pm {\scriptstyle 0.05 }$ & $ 0.94 \pm {\scriptstyle 0.05 }$ & $ 0.97 \pm {\scriptstyle 0.02 }$ & $ 0.91 \pm {\scriptstyle 0.07 } $ \\
& Healthy & $ \textbf{0.85} \pm {\scriptstyle 0.11 }$ & $ 0.83 \pm {\scriptstyle 0.09 }$ & $ \textbf{0.83} \pm {\scriptstyle 0.07 }$ & $ 0.95 \pm {\scriptstyle 0.03 }$ & $ 0.79 \pm {\scriptstyle 0.08 } $ \\
\hline \Tstrut \Bstrut
\multirow{3}{*}{\pbox{20cm}{\textbf{VGG-CAM}\\  \small{Acc.: 0.90, Bal.: 0.88} \\ \# Par.: 14 716 227}
} & COVID-19 & $ 0.93 \pm {\scriptstyle 0.05 }$ & $ 0.87 \pm {\scriptstyle 0.07 }$ & $ 0.90 \pm {\scriptstyle 0.05 }$ & $ 0.87 \pm {\scriptstyle 0.06 }$ & $ 0.79 \pm {\scriptstyle 0.09 } $ \\
& Pneumonia & $ 0.94 \pm {\scriptstyle 0.05 }$ & $ 0.95 \pm {\scriptstyle 0.05 }$ & $ 0.94 \pm {\scriptstyle 0.04 }$ & $ 0.98 \pm {\scriptstyle 0.02 }$ & $ 0.92 \pm {\scriptstyle 0.06 } $ \\
& Healthy & $ 0.78 \pm {\scriptstyle 0.10 }$ & $ 0.86 \pm {\scriptstyle 0.08 }$ & $ 0.81 \pm {\scriptstyle 0.05 }$ & $ 0.96 \pm {\scriptstyle 0.02 }$ & $ 0.77 \pm {\scriptstyle 0.06 } $ \\
\hline \Tstrut \Bstrut
\multirow{3}{*}{\pbox{20cm}{\textbf{NASNetMobile}\\  \small{Acc.: 0.76, Bal: 0.71} \\ \# Par.: 4 814 487}
}& COVID-19 & $ 0.87 \pm {\scriptstyle 0.10 }$ & $ 0.74 \pm {\scriptstyle 0.12 }$ & $ 0.80 \pm {\scriptstyle 0.10 }$ & $ 0.72 \pm {\scriptstyle 0.07 }$ & $ 0.59 \pm {\scriptstyle 0.15 } $ \\
& Pneumonia & $ 0.79 \pm {\scriptstyle 0.15 }$ & $ 0.88 \pm {\scriptstyle 0.08 }$ & $ 0.83 \pm {\scriptstyle 0.10 }$ & $ 0.96 \pm {\scriptstyle 0.03 }$ & $ 0.77 \pm {\scriptstyle 0.15 } $ \\
& Healthy & $ 0.47 \pm {\scriptstyle 0.03 }$ & $ 0.61 \pm {\scriptstyle 0.13 }$ & $ 0.53 \pm {\scriptstyle 0.05 }$ & $ 0.92 \pm {\scriptstyle 0.04 }$ & $ 0.43 \pm {\scriptstyle 0.07 } $ \\
\hline \Tstrut \Bstrut
\multirow{3}{*}{\pbox{20cm}{\textbf{VGG-Segment}\\  \small{Acc.: \textbf{0.91}, Bal: 0.89} \\ \# Par.: 34 018 074}}
& COVID-19 & $ \textbf{0.96} \pm {\scriptstyle 0.05 }$ & $ 0.89 \pm {\scriptstyle 0.06 }$ & $ \textbf{0.92} \pm {\scriptstyle 0.04 }$ & $ 0.88 \pm {\scriptstyle 0.07 }$ & $ \textbf{0.84} \pm {\scriptstyle 0.07 } $ \\
& Pneumonia & $ \textbf{0.96} \pm {\scriptstyle 0.03 }$ & $ \textbf{0.97} \pm {\scriptstyle 0.03 }$ & $ \textbf{0.96} \pm {\scriptstyle 0.02 }$ & $ \textbf{0.99} \pm {\scriptstyle 0.01 }$ & $ \textbf{0.95} \pm {\scriptstyle 0.03 } $ \\
& Healthy & $ 0.77 \pm {\scriptstyle 0.14 }$ & $ \textbf{0.91} \pm {\scriptstyle 0.08 }$ & $ 0.82 \pm {\scriptstyle 0.08 }$ & $ \textbf{0.97} \pm {\scriptstyle 0.03 }$ & $ 0.79 \pm {\scriptstyle 0.09 } $ \\
\hline \Tstrut \Bstrut
\multirow{3}{*}{\pbox{20cm}{\textbf{Segment-Enc}\\  \small{Acc.: 0.90, Bal: 0.89} \\ \# Par.: 19 993 307}}
& COVID-19 & $ 0.92 \pm {\scriptstyle 0.09 }$ & $ 0.91 \pm {\scriptstyle 0.06 }$ & $ 0.91 \pm {\scriptstyle 0.03 }$ & $ 0.90 \pm {\scriptstyle 0.06 }$ & $ 0.82 \pm {\scriptstyle 0.04 } $ \\
& Pneumonia & $ 0.95 \pm {\scriptstyle 0.04 }$ & $ 0.89 \pm {\scriptstyle 0.12 }$ & $ 0.92 \pm {\scriptstyle 0.07 }$ & $ 0.95 \pm {\scriptstyle 0.06 }$ & $ 0.89 \pm {\scriptstyle 0.08 } $ \\
& Healthy & $ 0.79 \pm {\scriptstyle 0.17 }$ & $ 0.89 \pm {\scriptstyle 0.10 }$ & $ 0.82 \pm {\scriptstyle 0.11 }$ & $ 0.98 \pm {\scriptstyle 0.02 }$ & $ \textbf{0.80} \pm {\scriptstyle 0.12 } $ \\
\bottomrule
\end{tabular}}
\caption{\textbf{Performance comparison.} Comparison of the tested classification models on 5-fold cross validation for each class. Acc. abbreviates accuracy, Bal. balanced accuracy, MCC Matthews Correlation Coefficient and Par. the number of parameters. 
For each class and each column the best model is highlighted in bold.
While \texttt{VGG} and \texttt{VGG-CAM} achieve an accuracy of $90\%$, the performance can, for the price of doubled computational costs, be further pushed with pre-trained lung US segmentation models from~\citet{roy2020deep}.
\texttt{NASNetMobile} instead is inferior but significantly smaller.
}
\label{tab:resultsclasses}
\end{table*}

\begin{figure}[t]
    \begin{subfigure}{0.24\textwidth}
        \includegraphics[width=\textwidth]{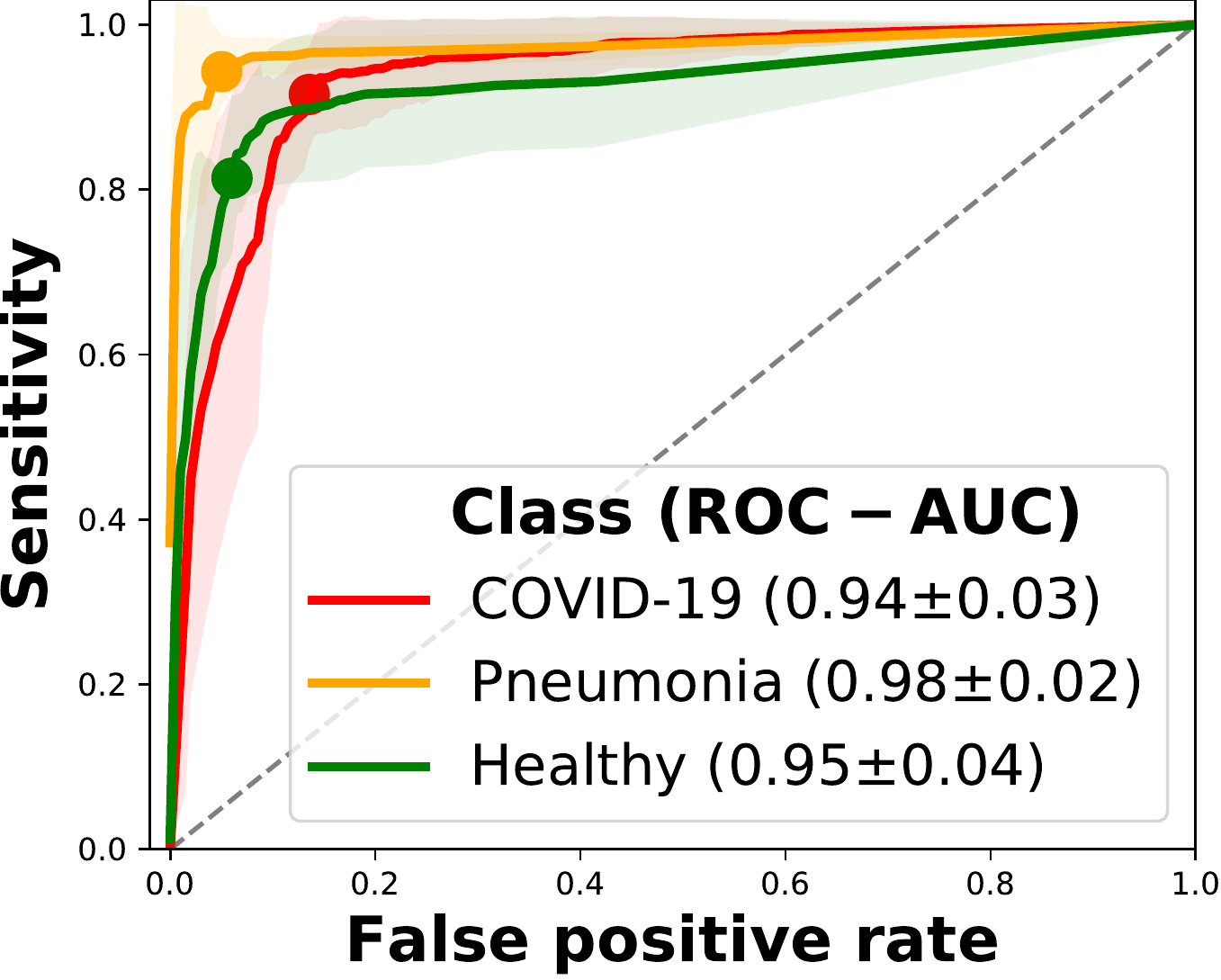}
        \caption{ROC-curves}
        \label{fig:roc}
    \end{subfigure}
    \hfill
    \begin{subfigure}{0.24\textwidth}
        \includegraphics[width=\textwidth]{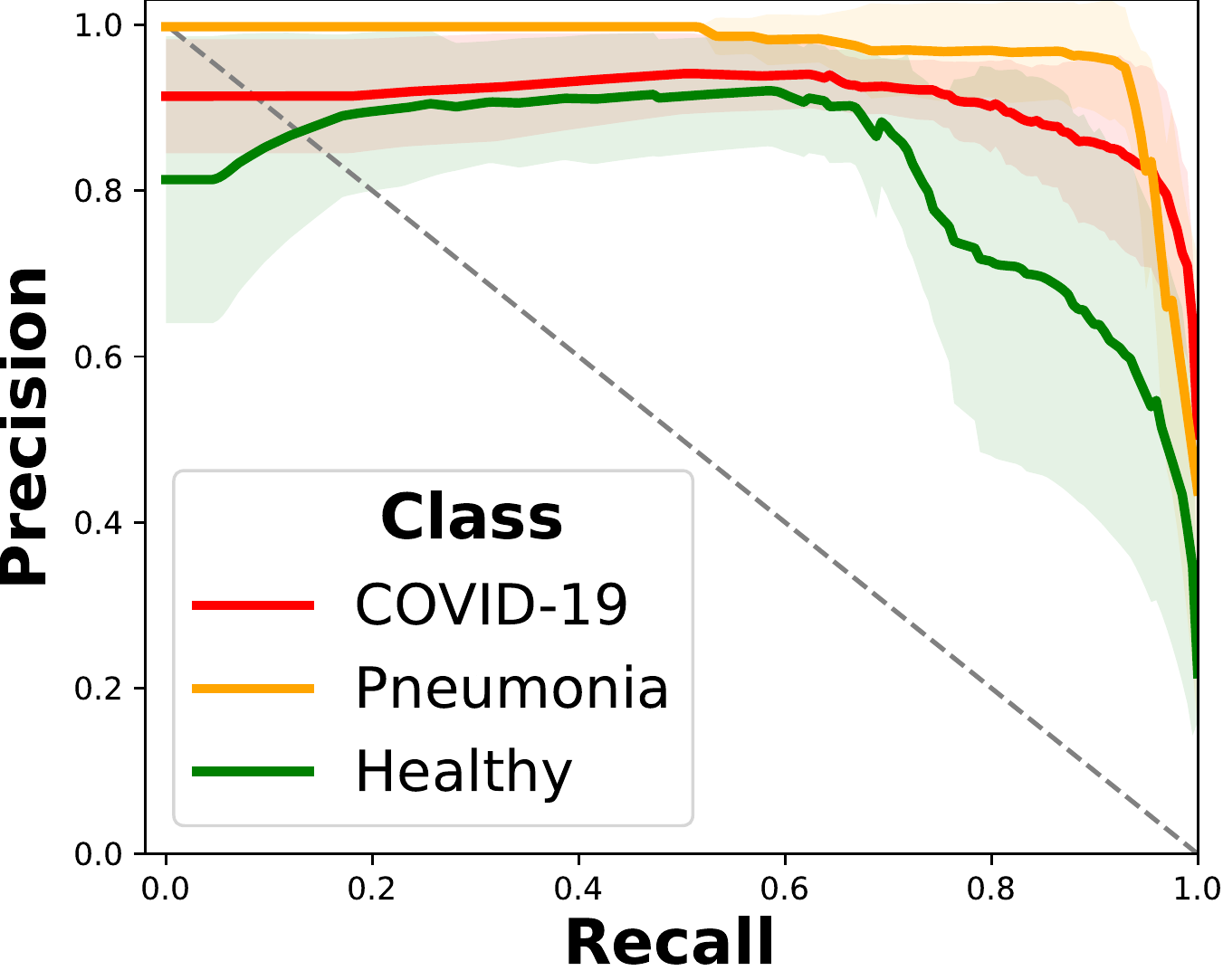}
        \caption{Pecision-recall-curves}
        \label{fig:precrec}
    \end{subfigure}
    \hfill
    \begin{subfigure}{0.24\textwidth}
        \includegraphics[width=\textwidth]{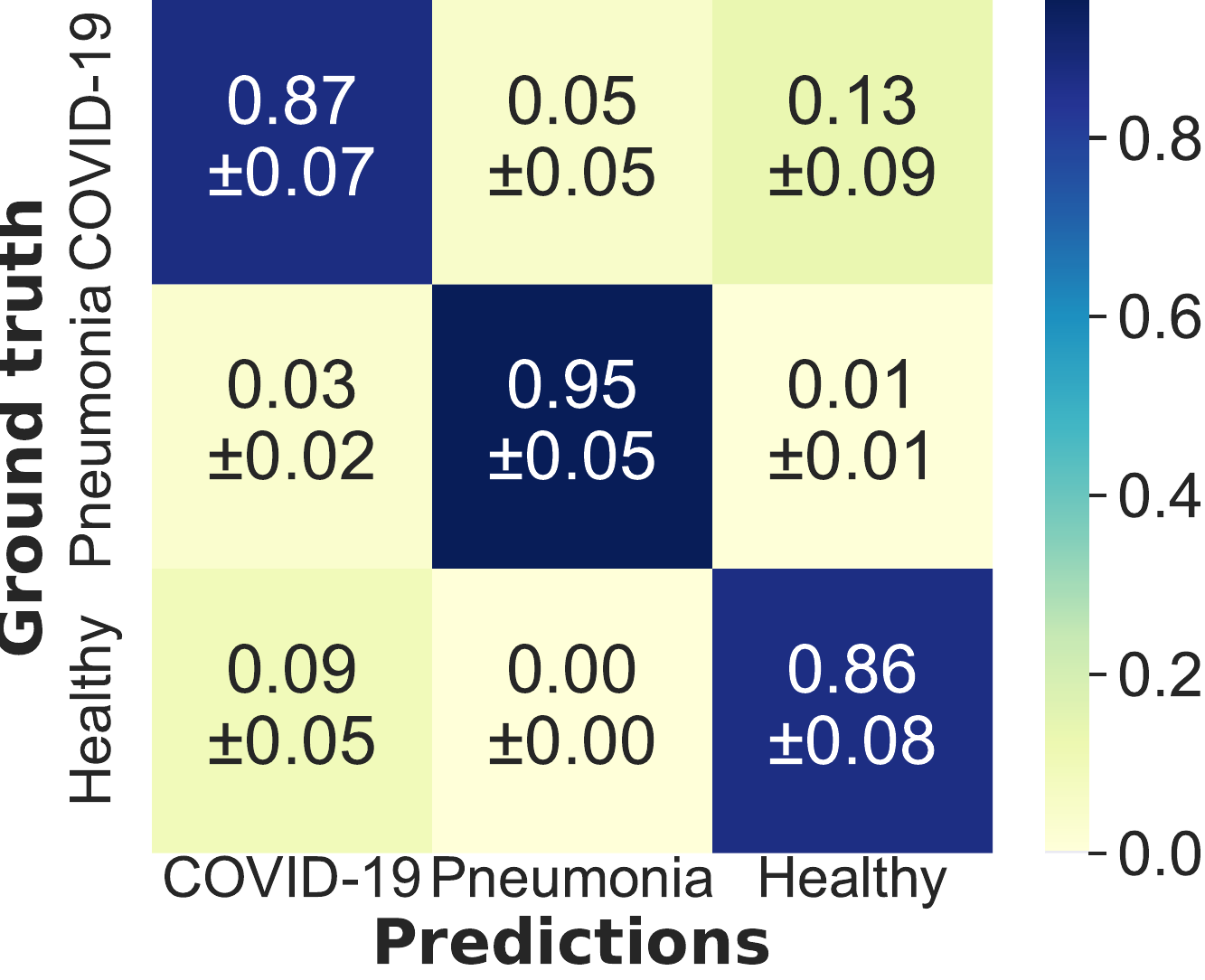}
        \caption{Precision-confusion matrix}
        \label{fig:precconf}
    \end{subfigure}
    \hfill
    \begin{subfigure}{0.24\textwidth}
        \includegraphics[width=\textwidth]{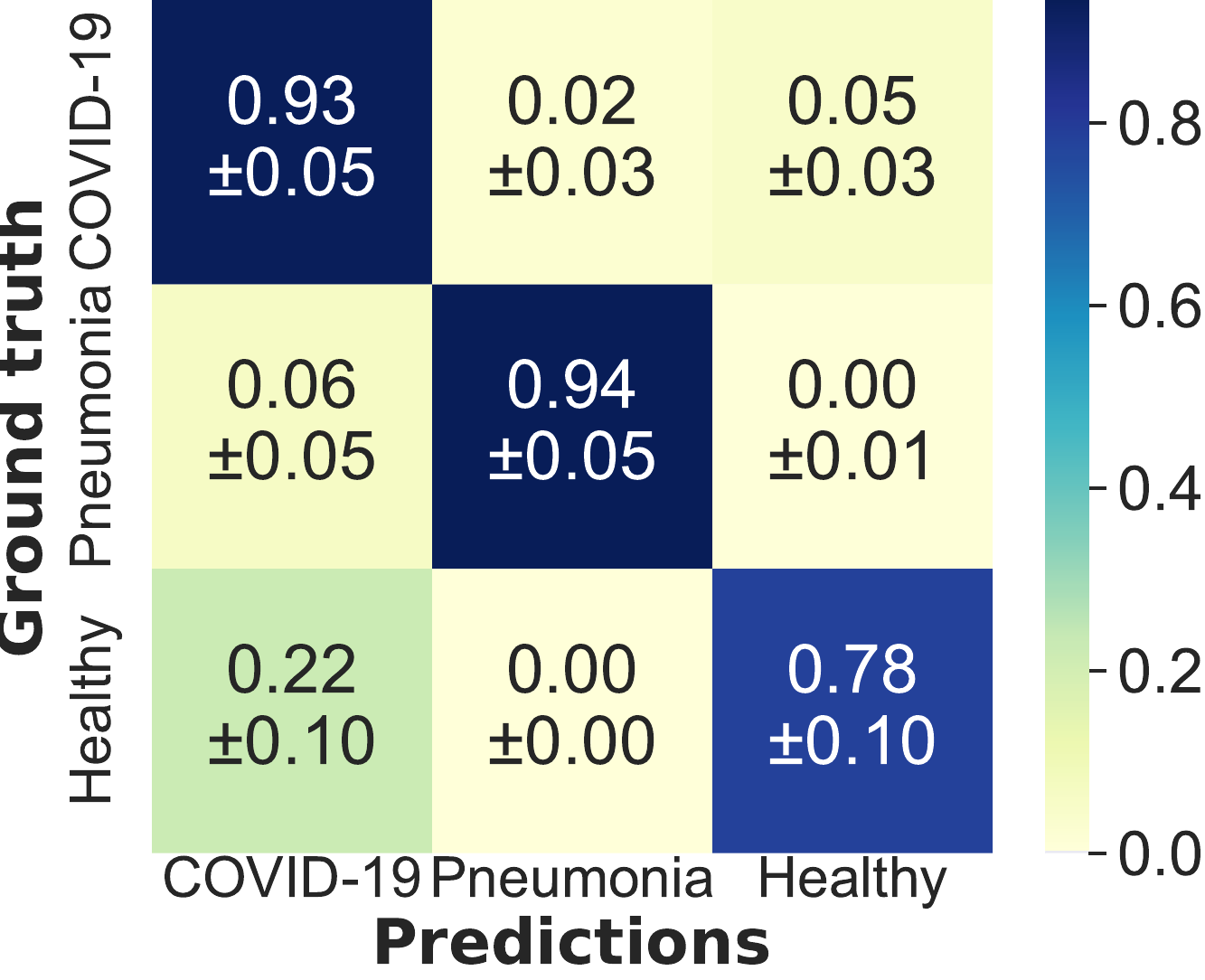}
        \caption{Sensitivity-confusion matrix}
        \label{fig:sensconf}
    \end{subfigure}
    \caption{\textbf{Performance of the \texttt{VGG-CAM} model}. Per-class ROC-AUC, sensitivity, and precision are shown on the diagonals of the normalized confusion matrices, highlighting the model's ability to distinguish COVID-19 from pneumonia and healthy lung images.
    }
    \label{fig:results}
\end{figure}

\autoref{tab:resultsclasses} shows a detailed comparison of the three best models in terms of recall, precision, specificity and F1-scores, as well as MCC.
Overall, both \texttt{VGG} and \texttt{VGG-CAM} achieve promising performance with an accuracy of $90\pm 2\%$ and $90\pm 5\%$ respectively on a 5-fold CV of 1,365 frames. 
Concerning per-class prediction accuracies, it is evident that bacterial pneumonia infections are distinguished best, with recall, precision, and specificity above $0.93$ for \texttt{VGG} and \texttt{VGG-CAM}, indicating the models' ability to recognize strong irregularities in lung images.
Although \texttt{VGG} slightly outperforms \texttt{VGG-CAM}, we explored the latter more in detail, due to its higher sensitivity for COVID-19 and its better performance when taking into account the class activation maps.
\autoref{fig:roc} visualizes the results of the~\texttt{VGG-CAM} model for each binary detection task as a ROC curve, showing ROC-AUC scores of $0.94$ and above for COVID-19 and the other two classes, while depicting the point where the accuracy is maximal for each class.
The false positive rate at the maximal-accuracy point is larger for COVID-19 than for pneumonia and healthy patients.
In a clinical setting, where false positives are less problematic than false negatives, this property is highly desirable.
Since the data is imbalanced, we also plot the precision--recall curve in \autoref{fig:precrec}, which confirms that pneumonia is the class that is predicted most easily. %~\citep{davis2006relationship}.
In addition, the confusion matrices in \autoref{fig:precconf} and \autoref{fig:sensconf} further detail the predictions of \texttt{VGG-CAM}; we observe that the high sensitivity for COVID-19~($0.93$, $642$ out of $693$ frames) comes at a cost of $22\%$ false positives from the healthy class.
%Further work is necessary to improve the false positive rate of COVID-19, as 64 healthy-lung images and 23 pneumonia-infected lungs are classified as COVID-19.
For further results including the ROC- and precision--recall curves of all three models see Appendix~\ref{appendix:results}.

\paragraph{Ablation study with segmentation models.}
Lung US recordings are noisy and operator-dependent, posing difficulties for the classification of raw data.
Hence, we compare \texttt{VGG} and \texttt{VGG-CAM} to \texttt{VGG-Segment} where all frames are segmented~(i.e.\ classified on a pixel level into pathological patterns) with the model from~\citet{roy2020deep}; see Appendix~\ref{appendix:arch}) for an example input.
The relevant rows in \autoref{tab:resultsclasses} exhibit mixed results: while training on segmented images \emph{improves} most relevant performance metrics slightly~(higher accuracy, COVID-19 sensitivity, and MCC scores), balanced accuracy is \emph{inferior} compared to \texttt{VGG}.
Since this small increase in predictive performance comes at the cost of a large increase in model size~(due to the ensemble of three independent models; selecting only one of the models resulted in inferior performance), we considered \texttt{Segment-Enc}, i.e.\ a dense model classifying the $560$-dimensional encoding produced by the pre-trained segmentation models. 
\texttt{Segment-Enc} achieved comparable performance for most metrics, apart from lower scores for pneumonia detection. 
Since the difference in performance is only marginal, and the architectures of \texttt{VGG-Segment} and \texttt{Segment-Enc} prohibit the computation of class activation maps, we prefer to focus on the analysis of \texttt{VGG-CAM} in the following.

\paragraph{Ablation study on other architectures.}
Initially, we had tested further models proposed for medical image analysis, such as \texttt{COVID-Net}~(previously used for the classification of X-Ray images~\cite{wang2020covid}), and an architecture following \cite{li2020artificial} based on a \texttt{Res-Net}~\cite{he2016deep}, but we observed that the experiments on our data resulted in \emph{significantly worse} results.
Last, we tested several smaller networks such as \texttt{MobileNet}\cite{howard2017mobilenets} as an additional ablation study, with \texttt{NASNetMobile}~\citep{zoph2018learning} performing best. 
As most ultrasound devices are portable and real-time inference on the devices is technically feasible, 
resource-efficient networks are highly relevant and could supersede web-based inference.
Due to low precision and recall on healthy data, our fine-tuned \texttt{NASNetMobile} is less performant than \texttt{VGG-CAM}, but also requires less than a third of the parameters, thus providing a first step towards real-time on-device inference.

\subsection{Video-based experiments}

To investigate the need for a model with the ability to detect spatiotemporal patterns in lung US, we explored \texttt{Models Genesis}, a pretrained 3D-CNN designed for 3D medical image analysis~\citep{zhou2019models}.
%In an application setting the input data is usually a video. 
%Considering that the medical experts who manually checked our dataset report difficulties in judging from single images, future work should study video classification techniques instead of frame-based methods.
%Here, we present a first attempt for video classification employing a pre-trained 3D-CNN called Models Genesis~\citep{zhou2019models}.
%
%
%
%\paragraph{Results.}
\autoref{tab:resultsvideo} contrasts the frame-based performance of \texttt{VGG-CAM} model to \texttt{Model Genesis}. The video classifier is outperformed by \texttt{VGG-CAM}, with a video accuracy of 94\% compared to 87\%.
Note that all videos of pneumonia-infections are classified correctly, while especially \texttt{Model Genesis} struggles with the prediction of healthy patients. Considering that only 292 video-chunks were available for training \texttt{Model Genesis}, while 1356 images are used to train \texttt{VGG-CAM}, even extended through data augmentation techniques, it is likely that video-based classification may improve with increasing data availability.

% \input{video_class_table}

% Table spreading both columns
\begin{table}[t]
  \centering
 \scalebox{0.9}{
\begin{tabular}{llccccc}
\toprule
{} &    \textbf{Class} & \textbf{Recall} &  \textbf{Precision} &   \textbf{F1-score} &     \textbf{Specificity} &   \textbf{MCC}  \\
\midrule  \Tstrut \Bstrut
\multirow{3}{*}{\pbox{20cm}{\textbf{VGG-CAM}\\  \small{Acc.: \textbf{0.94}, Bal.: \textbf{0.93}} \\ \# Par.: 14, 716 227}} % ACC 0.94 STD 0.04 BAL 0.93 STD 0.05
& COVID-19 & $ \textbf{0.98} \pm {\scriptstyle 0.04 }$ & $ \textbf{0.91} \pm {\scriptstyle 0.08 }$ & $ \textbf{0.94} \pm {\scriptstyle 0.04 }$ & $ 0.91 \pm {\scriptstyle 0.08 }$ & $ \textbf{0.89} \pm {\scriptstyle 0.06 } $ \\
& Pneumonia & $ \textbf{1.00} \pm {\scriptstyle 0.00 }$ & $ \textbf{1.00} \pm {\scriptstyle 0.00 }$ & $ \textbf{1.00} \pm {\scriptstyle 0.00 }$ & $ \textbf{1.00} \pm {\scriptstyle 0.00 }$ & $ \textbf{1.00} \pm {\scriptstyle 0.00 } $ \\
& Healthy & $ 0.80 \pm {\scriptstyle 0.16 }$ & $ \textbf{0.96} \pm {\scriptstyle 0.08 }$ & $ \textbf{0.86} \pm {\scriptstyle 0.08 }$ & $ \textbf{0.98} \pm {\scriptstyle 0.03 }$ & $ \textbf{0.84} \pm {\scriptstyle 0.09 } $ \\
\hline  \Tstrut \Bstrut
\multirow{3}{*}{\pbox{20cm}{\textbf{Models Genesis}\\  \small{Acc.: 0.87, Bal.: 0.87} \\ \# Par.: 7,559,043}
}
% ACC 0.87 STD 0.05 BAL 0.87 STD 0.05
& COVID-19 & $ 0.80 \pm {\scriptstyle 0.19 }$ & $ 0.90 \pm {\scriptstyle 0.09 }$ & $ 0.83 \pm {\scriptstyle 0.11 }$ & $ \textbf{0.92} \pm {\scriptstyle 0.08 }$ & $ 0.74 \pm {\scriptstyle 0.11 } $ \\
& Pneumonia & $ \textbf{1.00} \pm {\scriptstyle 0.00 }$ & $ \textbf{1.00} \pm {\scriptstyle 0.00 }$ & $ \textbf{1.00} \pm {\scriptstyle 0.00 }$ & $ \textbf{1.00} \pm {\scriptstyle 0.00 }$ & $ \textbf{1.00} \pm {\scriptstyle 0.00 } $ \\
& Healthy & $ \textbf{0.82} \pm {\scriptstyle 0.15 }$ & $ 0.75 \pm {\scriptstyle 0.21 }$ & $ 0.75 \pm {\scriptstyle 0.07 }$ & $ 0.89 \pm {\scriptstyle 0.10 }$ & $ 0.69 \pm {\scriptstyle 0.09 } $ \\
\bottomrule
\end{tabular}
}
\caption{\textbf{Video classification results.} The frame-based model \texttt{VGG-CAM} outperforms the 3D CNN \texttt{Models Genesis}, showing high accuracy (94\%), recall, precision for COVID-19 and pneumonia detection.}
\label{tab:resultsvideo}
\end{table}

\subsection{Evaluation on independent test data}
Very recently, the \texttt{ICLUS} initiative released $60$ COVID-19 lung US recordings from Italian patients\footnote{$40$ convex + $20$ linear probes are available from~\url{https://iclus-web.bluetensor.ai/}}\citep{roy2020deep}.
The data is not annotated, but was initially assumed to contain only COVID-19 videos, based on its general description.
We evaluated the performance of the~\texttt{VGG-CAM} model on all $40$ convex probes from \texttt{ICLUS}, alongside $24$ recordings from healthy controls~($6$ viewpoints each) and $2$ videos from public sources~(healthy), jointly comprising an independent test dataset of $66$ videos. 
%Very recently we were provided further healthy-patient data of 4 volunteers (6 videos from different viewpoints each, self-recorded), as well as two other videos of normal lungs found online.
%In addition, ultrasound data of COVID-19 patients became available in the context of the ICLUS initiative~\citep{roy2020deep}, including 40 videos of convex probes. This data thus comprises an independent test set of 40 COVID-19 and 26 healthy-patient data. 

We predicted all frames as an average of the five \texttt{VGG-CAM} models trained in cross-validation. The model achieves a frame-prediction accuracy of $83.3\%$, divided into $89.5\%$ for healthy-patient data and $74\%$ for COVID-19 videos.
Furthermore, averaging the class probabilities over all frames, \texttt{VGG-CAM} achieves a video classification accuracy of $92.2\%$ and $77.5\%$, respectively.
Notably, the four healthy patients are \emph{all} classified correctly if summarized across viewpoints. Combining both datasets, the sensitivity of detecting COVID-19 corresponds to the accuracy~(0.775) with a precision score of 0.94~(no video was classified as bacterial pneumonia). 

\paragraph{Evaluation by domain experts.}
We further investigated the comparably low sensitivity on the COVID-19 data~(\texttt{ICLUS}) with the help of two medical experts.
When asked for their unbiased diagnosis of the incorrectly-predicted videos, they independently reported for $6$ out of $9$ videos that no disease-specific patterns can be observed~(``A-lines, normal lung''). While these findings support the performance of our model, the \emph{true} label of the data remains unclear.
In addition, the dataset may contain further healthy-patient data which was incorrectly predicted as COVID-19.
At this point, we can safely conclude that test data performance is highly promising, in particular considering the high accuracy for healthy patients, but requires further validation with independent and labeled data. 

% Patient 2 asthma

\section{Model explainability}

\subsection{Class activation maps}

\begin{figure}[tb]
\includegraphics[width=\linewidth]{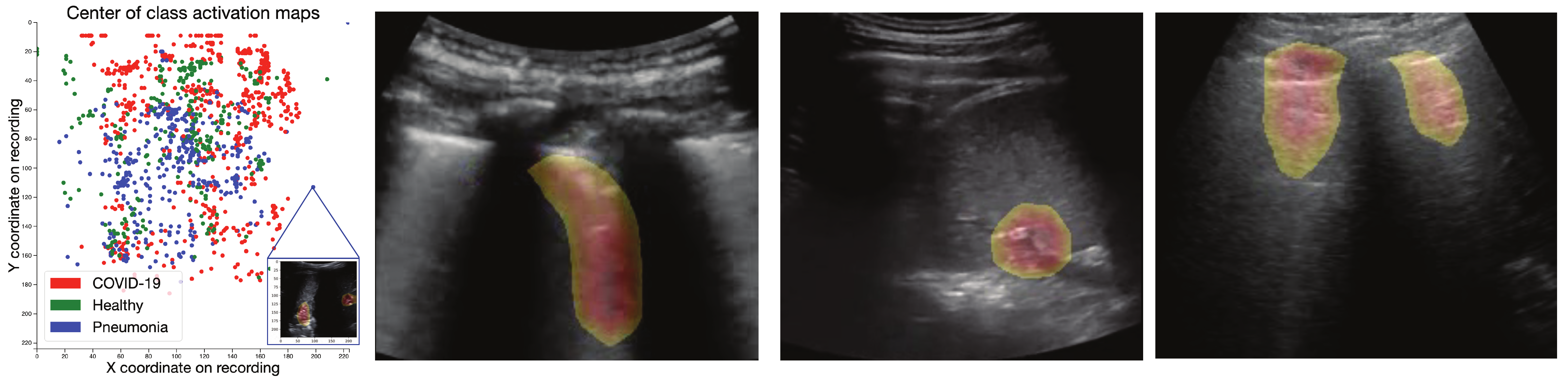}
\caption{
\textbf{Class activation maps.}
\textbf{Left:} Interactive scatterplot of the origins of the CAMs across the entire dataset, colored by class.
While the data seems rather unstructured, pneumonia-CAMs have lower $y$-coordinates than COVID-19 and healthy samples.
\textbf{Rest:} Exemplary CAMs for COVID-19 (highlighting a B-line), bacterial pneumonia (highlighting pleural consolidations) and healthy lungs (highlighting A-lines).
}
\label{fig:cams}
\end{figure}

Class activation mapping (CAM) are a popular technique for model explainability that exploits global average pooling and allows to compute class-specific heatmaps that indicate the discriminative regions of the image that caused the particular class activity of interest~\citep{zhou2016learning}.
For healthcare applications, CAMs, or their generalization Grad-CAMs~\citep{selvaraju2017grad}, can provide valuable decision support by unravelling whether a model's prediction was based on visible pathological patterns. 
Moreover, CAMs can guide doctors and point to informative patterns, especially relevant in time-sensitive~(triage) or knowledge-sensitive~(third-world countries) situations.

\paragraph{Results.}
\autoref{fig:cams} shows representative CAMs in the three rightmost panels. 
They highlight the most frequent US pattern for the three classes, COVID-19~(vertical B-lines), bacterial pneumonia~(consolidations), and healthy~(horizontal A-line).
For a more quantitative estimate, we computed the points of maximal activation of the CAMs for each class~(abbreviated as $\mathbf{C}$, $\mathbf{P}$, and $\mathbf{H}$) and all samples of the dataset~(see~\autoref{fig:cams} left).
While, in general, the heatmaps are fairly distributed across the probe, pneumonia related features were rather found in the center and bottom part, especially compared to COVID-19 and healthy patterns\footnote{The interactive HTML and a few exemplary CAM videos are available at: \url{https://bit.ly/3eASPc8}}.
Please refer to Appendix~\ref{appendix:cams} for a density plot.
To assess to what extent the differences between the individual distributions are significant, we employed \emph{maximum mean discrepancy}~(MMD), a metric between statistical distributions~\citep{gretton2012kernel} that enables the comparison of distributions via kernels, i.e.\ generic similarity functions. Given two coordinates $x,y\in\reals^2$ and a smoothing parameter $\sigma\in\reals$, we use a Gaussian kernel $k(x, y) := \exp(-\nicefrac{\|x - y\|^2}{\sigma^2})$ to assess the dissimilarity between $x$ and $y$. Following \citet{gretton2012kernel}, we set $\sigma$ to the median distance in the aggregated samples~(i.e.\ all samples, without considering labels). We then calculate MMD values for the distance between the three classes, i.e.\ $\MMD(\mathbf{C}, \mathbf{P}) \approx 0.0051$, 
$\MMD(\mathbf{C}, \mathbf{H}) \approx 0.0061$, 
and
$\MMD(\mathbf{P}, \mathbf{H}) \approx 0.0065$.
Repeating this calculation for $5000$ bootstrap samples per class~(see \autoref{fig:MMD bootstrap histograms} for the resulting histograms), we find that the observe achieved significance levels of the intra-class MMD values of well below an $\alpha = 0.05$ significance level.
 
\paragraph{Expert validation of CAMs for Human-in-the-loop settings.}
A potential application of our framework is a human-in-the-loop (HITL) setting with CAMs as a core component of the decision support tool that highlights pulmonary biomarkers and guides the decision makers.
Since the performance of qualitative methods like CAMs can only be validated with the help of doctors, we conducted a blind-folded study with two medical experts experienced in the diagnostic process with ultrasound recordings.
The experts were shown $50$ videos~($14$ COVID-19, $21$ pneumonia, $14$ regular) comprising all non-proprietary video data which was correctly classified by the model.
The class activation map for the respective class was computed two times, first with an average of all five models that were trained, and second only with the model that did not see any frame of the video during training~(called train- and test-CAMs in the following). Both experts were asked to compare both activation maps for all $50$ videos, and to score them on a scale of $-3$ (``the heatmap is only distracting'') to $3$~(``the heatmap is very helpful for diagnosis''). 

\begin{figure}[!htb]
    \vspace{-4mm}
    \centering
    \includegraphics[width=0.5\textwidth]{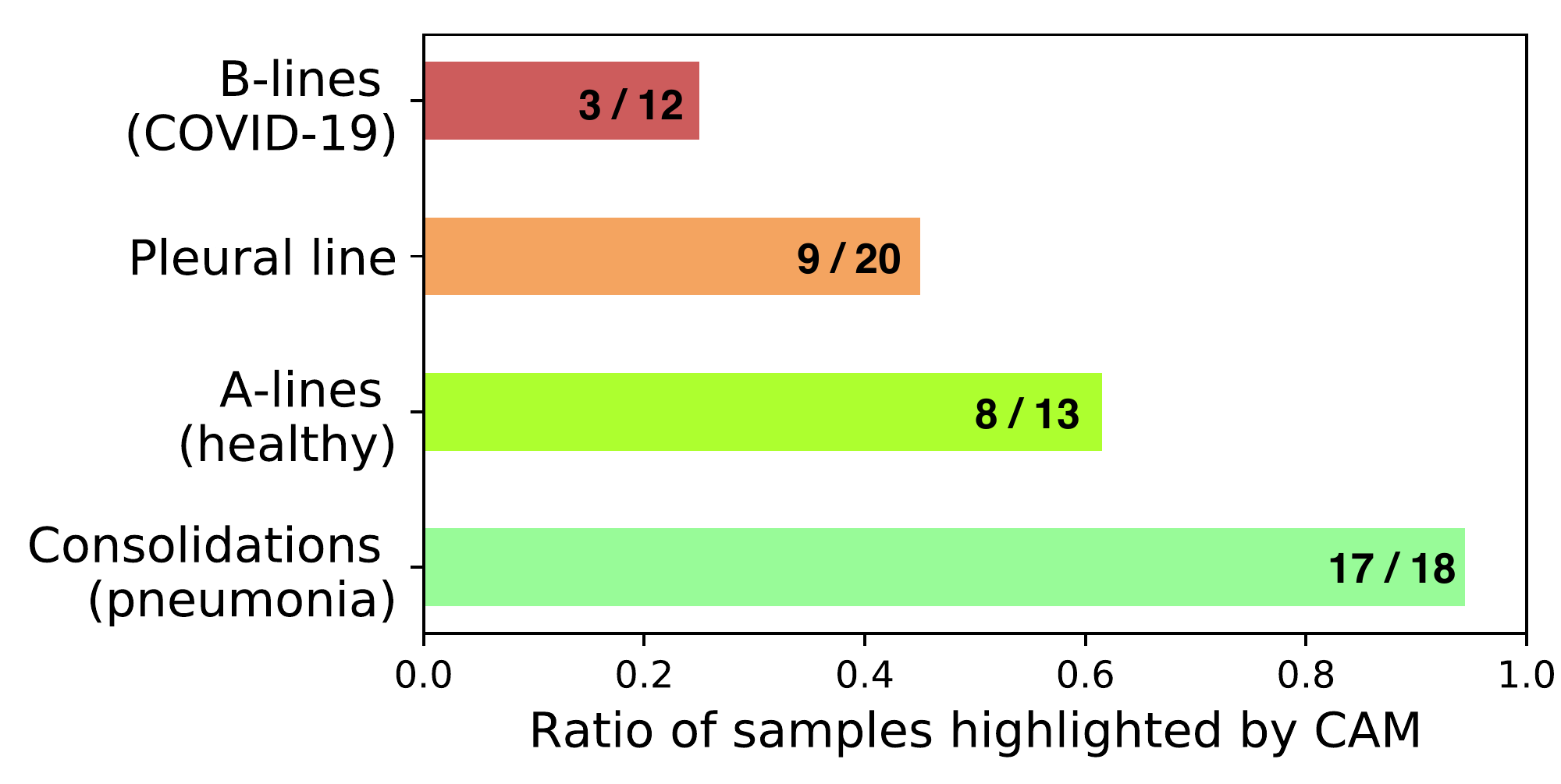}
    \caption{Patterns highlighted by CAMs compared to visible patterns in the video.}
    \label{fig:patterns}
\end{figure}

First, the CAMs were overall perceived useful and the train and test CAMs were assigned a \textit{higher} average score of $0.45$ and $0.81$ respectively.
Second, disagreeing in only $8\%$ of the cases, both experts independently decided for the test-CAM with a probability of $56\%$.
Hence, the test-CAMs are not inferior to the
train-CAMs, however non-significant in a Wilcoxon signed-rank test.
% \todo{Add t-test for train vs.  test heatmap - NW: not significant see above}
However, train- and test-CAM both scored best for videos of bacterial pneumonia, lacking performance for videos of healthy and COVID-19 patients.
Specifically, test-CAM received an average score of $0.81$, divided into $-0.25$ for COVID-19, $2.05$ for pneumonia, and $0$ %-0.21 and -0.28 for COVID-19, 1.67 and 2.43 for pneumonia and -0.36 and 0.36
for healthy patients.
Third, the experts were asked to name the pathological patterns visible in general, as well as the patterns that were highlighted by the heatmap.
\autoref{fig:patterns} shows the average ratio of pattern that were correctly highlighted by the CAM model, where the patterns listed by the more senior expert are taken as the ground truth for each video.

% \begin{wrapfigure}[7]s{r}{0.4\textwidth}
%     \vspace{-4mm}
%     \centering
%     \includegraphics[width = 0.4\textwidth]{figures/barplot_cam.pdf}
%     \caption{Patterns highlighted by CAMs compared to visible patterns in the video.}
%     \label{fig:patterns}
% \end{wrapfigure}

Interestingly, the high performance of our model in classifying videos of bacterial pneumonia is probably explained by the model's ability to detect consolidated areas, where $17$ out of $18$ are correctly classified.
Moreover, A-lines are highlighted in $\sim60$\% of the normal lung recordings.
Problematically, in $13$ videos mostly fat, muscles or skin is highlighted, which has to be studied and improved in future work.
%

%\begin{wrapfigure}[8]{h}[0cm]{0.4\textwidth}\centering
%includegraphics[width=0.35\textwidth]{figures/barplot_cam.pdf}
%\caption{dheiua}
%\end{wrapfigure}
%
% In summary, \todo{shortly summarize results} % \texttt{POCOVID-Net} is able to learn frame-wise classification into COVID-19, pneumonia and healthy, where sensitivity for COVID-19 is already very high at 96\%.
% It was demonstrated that the proposed architecture outperforms \texttt{COVID-Net}, and aggregating the frame-wise predictions into video classification yields a general classification performance of 92\% accuracy. Last, it was argued that further work is required to reduce the number of false positive predictions of healthy lungs as COVID-19.

\subsection{Confidence estimates}
The ability to quantify states of high uncertainty is of crucial importance for medical image analysis and computer vision applications in healthcare. 
We assessed this via independent measures of epistemic~(model) uncertainty~(by drawing Monte Carlo samples from the approximate predictive posterior~\citep{gal2016dropout}) and aleatoric~(data) uncertainty (by means of test time data augmentation~\citep{ayhan2018test}). 
The sample standard deviation of 10 forward passes is interpreted as inverse, empirical confidence score $\in [0,1]$~(for details see appendix).
The epistemic confidence estimate was found to be highly correlated with the correctness of the predictions~($\rho=0.41, p<4\mathrm{e-}124$, mean confidence of $0.75$ and $0.26$ for correct and wrong predictions), while the aleatoric confidence was found correlated to a lesser extent ($\rho=0.29, p<6\mathrm{e-}35$, mean confidence of $0.88$ and $0.73$, respectively).
Across the entire dataset, both scores are highly correlated ($\rho=0.52$), suggesting to exploit them jointly to detect and remove predictions of low confidence in a possible application.

\section{Discussion}
Ultrasound as an established diagnosis tool that is both safe and highly available constitutes a method with potentially huge impact that has nevertheless been neglected by the machine learning community.
% \todo{ultrasound dependent on examiner} - was put in statement of impact
% \todo{Add that we are the first to demonstrate - Done in paragraph below}
% Using this data, we are the first to demonstrate feasibility of automatic detection of COVID-19 and bacterial pneumonia from ultrasound.
This work presents methods and analyses that pave the way towards computer vision-assisted differential diagnosis of COVID-19 from US,
% the first approach for automatic differential diagnosis of COVID-19 from US, 
providing an extensive analysis of (interpretable) methods that are relevant not only in the context of COVID-19, but in general for the diagnosis of viral and bacterial pneumonia.
% In the current global pandemic, it is as relevant as hardly ever before that the research community pools its expertise interdisciplinary to provide solutions in the near future.
% Our contribution is the exploration of automatizing COVID-19 detection from lung ultrasound imaging in order to provide a quick assessment of the possibility of a person to be infected with COVID-19.
 
% Our first step towards this goal is to release a collection of POCUS images and videos, that were gathered and pre-processed from the referenced sources.
% The videos are reliably labeled and can be split to generate a dataset of more than a thousand images.
% We would like to invite researchers to contribute to our database, e.g by pointing to new publications or by making lung ultrasound recordings available.
% We will constantly update this dataset on:~\url{https://github.com/jannisborn/covid19_pocus_ultrasound}.
We provide strong evidence that automatic detection of COVID-19 is a promising future endeavour and competitive compared to CT and CXR based models, with a sensitivity of 98\% and a specificity of 91\% for COVID-19, achieved on our dataset of 106 lung US videos.
In comparison, sensitivity up to 98\% and specificity up to 92\% was reported for CT \cite{butt2020deep, mei2020artificial}.
%
% Cohen et. al. [51] who started building anopen-access database of X-ray (now also CT images) that,to date, contains∼150 COVID-19 images:  Using deepconvolutional neural networks, many have claimed strongperformances on the X-ray data of Cohen et al, rangingfrom 91% up to 98%
% \todo{Add sensitivity/specificty on ICLUS/independent test data }
%Our analysis yields results comparable to automatic detection systems for CT or X-ray data. 
% \todo{Put our performance numbers in perspective to CT/XCR models} 
We verified our results with independent test data, studied model uncertainty and concluded a significant ability of our model to recognize low-confidence situations.
We combined our approach with the only available related work, lung US segmentation models from~\citet{roy2020deep}, and found mild performance improvement in most metrics.
It however remains unclear whether this gain can be attributed to the segmentation itself or is a side-effect of the increased parametrization.
Certainly, there are many approaches yet to be explored in order to improve on the results presented here, including further work on video classification, but also exploiting the higher availability of CT or X-ray scans with transfer learning or adapting generative models to complement the scarce data about COVID-19 as proposed in~\cite{loey2020within}.
% From the machine learning perspectives several improvements to \texttt{POCOVID-Net} are possible and should be considered, given more data becomes available. First, the benefit of pre-training the network on large image databases could be improved by training the model on (non lung) ultrasound samples instead of using \texttt{ImageNet}, a database of real life objects. This pre-training may help detecting ultrasound specific patterns such as B-Lines. 
Furthermore, we investigated the value of interpretable methods in a quantitative manner with the implementation and validation of class activation mapping in a study involving medical experts. 
% Interpratable methods are highly relevant for the value of development of a decision support tool. 
While the analysis provides excellent evidence for the successful detection of pathological patterns like consolidations, A-lines and effusion, it reveals problems in the model's ''focal point'' (e.g.  missing B-lines and sometimes highlighting muscles instead of the lung) which should be further addressed using ultrasound segmentation techniques~\citep{van2019localizing}.
% JB: I'd leave out other viral pneumonias, not rerlevant for this target group. NW: fine by me, it's in the data table though, we should mention it somewhere
%Another open question to address is the differentiation of COVID-19 from other viral pneumonia

% In short, we provide methods and analyses that pave the way towards computer vision-assisted differential diagnosis of COVID-19 from US.
%Lung ultrasound may not only take a key role in disease diagnosis, but can be utilized to monitor disease evolution through regular checks performed non-invasively and without the need of relocation.
Our published database is constantly updated and verified by medical experts researchers are invited to contribute to our initiative. 
%
% mention ablation study in conclusion - more application related
We envision the proposed tool as a decision support system to accelerate diagnosis or provide a ``second opinion'' to increase reliability.
We started to collaborate with radiologists and an intensive care unit and are currently designing a controlled, clinical study to investigate the predictive power of US for automatic detection of COVID-19, especially in comparison to CT and CXR.
As a preliminary demonstration, we have built a web service (link not anonymized) where users can screen ultrasound images, querying our averaged prediction model.
We aim to extend the functionality of the website in the future to offer interpretable video inference, aiming for an accessible and validated tool that enables medical doctors to draw inference from their US images with unprecedented ease, convenience and speed.

\bibliographystyle{abbrvnat}
\begin{small}
\bibliography{references}  %%% Remove comment to use the external .bib file (using bibtex).
\end{small}

\clearpage
% \input{appendix}

\begin{appendices}

% TAKEN OUT FOR DOUBLE BLIND
% \section*{Acknowledgements}
% \todo{Add pocovodscreen people}
% We would like to thank Ruud van Sloun for sharing pre-trained segmentation models.
%In particular, you should not include author names, author affiliations, or acknowledgements in your submission and you should avoid providing any other identifying information (even in the supplementary material)

% \counterwithin{figure}{section}
% \counterwithin{table}{section}
\begin{appendix}
\section{Appendix}

\subsection{Model architectures and hyperparameter}
\label{appendix:arch}
As a base, we use the convolutional part of the established \texttt{VGG-16}~\cite{Simonyan15}, pre-trained on \texttt{Imagenet}.
The model we call \texttt{VGG} is followed by one hidden layer of 64 neurons with \texttt{ReLU} activation, dropout of 0.5, batch normalization and the output layer with \texttt{softmax} activation. 
The CAMs for this model were computed with Grad-CAM~\citep{selvaraju2017grad}.
To compare Grad-CAMs with regular CAMs~\citep{zhou2016learning}, we also tested \texttt{VGG-CAM}, a CAM-compatible VGG with a single dense layer following the global average pooling after the last convolutional layer.
For both models, during training only the weights of the last three layers were fine-tuned, while the other ones were frozen to the values from pre-training.
%The CNN is then followed by two two fully connected layers of sizes 64 and three neurons respectively, where dropout (rate: 0.5) and batch normalization are applied in between.
This results in a total of $\sim2.4$M trainable and $\sim12.4$M non-trainable parameters.
The model is trained with a cross entropy loss function on the \texttt{softmax} outputs, and optimized with \texttt{Adam} with an initial learning rate of $1\mathrm{e}{-4}$.
All models were implemented in~\texttt{TensorFlow} and trained for 40 epochs with a batch size of 8 and early stopping was enabled.

\subsection{Pretrained segmentation models}\label{appendix:segment}

\begin{figure}[htb!]
\centering
        \includegraphics[width=0.8\textwidth]{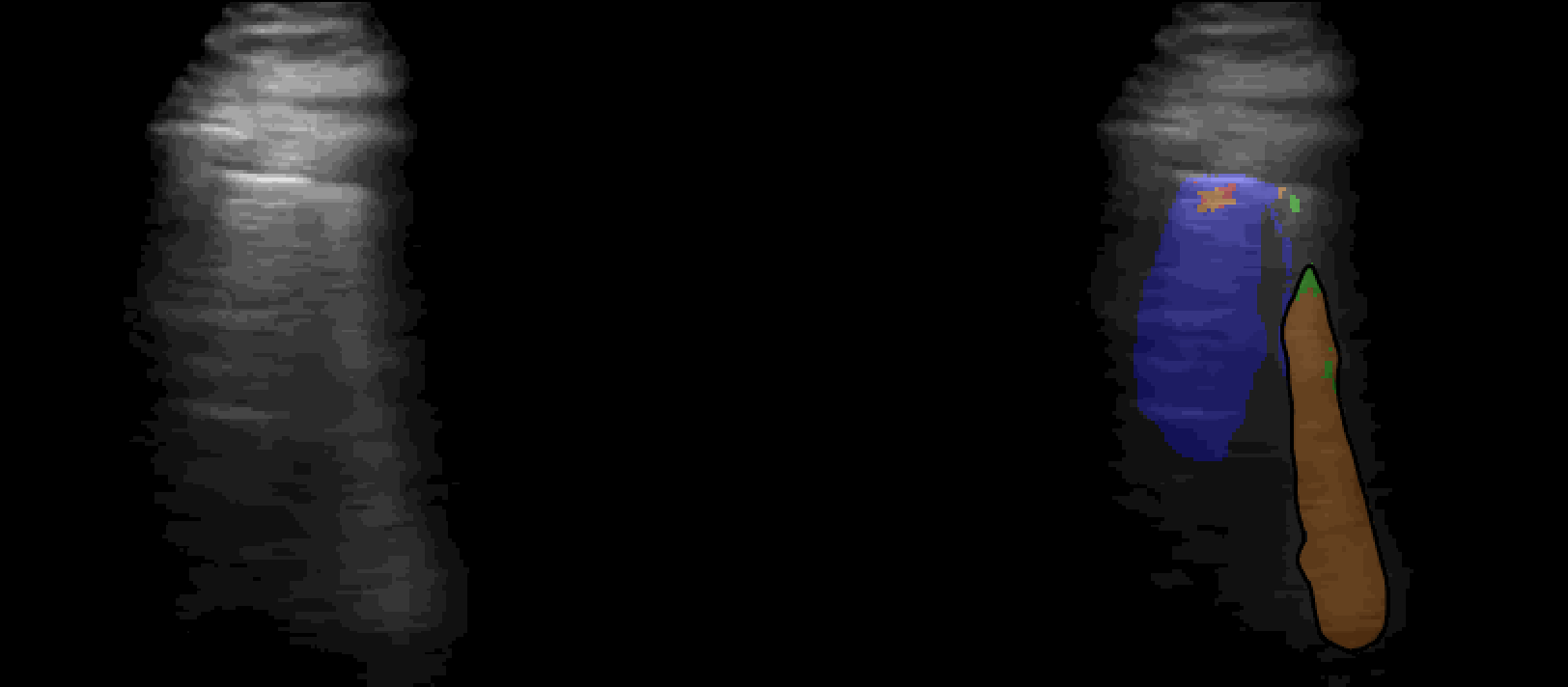}
    \caption{\textbf{Example snapshot from lung segmentation of COVID-19 patient.}
    Left side shows the raw US recording and the right side shows the segmentation method from~\citet{roy2020deep}
    highlighting the B-line. 
    The images shown on the right were used as input for the~\texttt{VGG-Segment} model.
    }
    \label{fig:segmented}
\end{figure}

\autoref{fig:segmented} gives an example for the segmented ultrasound image with the model from~\citet{roy2020deep}. In our work the segmented image serves as input to the \texttt{VGG-Segment} model.
\FloatBarrier

\subsection{Uncertainty estimation}
For both aleatoric and epistemic uncertainty, the confidence estimate $c_i$ of sample $i$ is computed by scaling the sample's standard deviation to $\in [0,1]$ and interpreting it as an inverse precision:
\begin{equation}
    \label{eq:confidence}
c_i = -(\frac{\sigma_{i,j} - \sigma_{min}}{\sigma_{max}-\sigma_{min}}) + 1  \, , 
\end{equation}
where $\sigma_{i,j}$ is the sample standard deviation of the ten class probabilities of the winning class $j$, $\sigma_{min}$ is the minimal standard deviation (0, i.e. all probabilities for the winning class are identical) and $\sigma_{max}$ is the maximal standard deviation, i.e. 0.5.
Practically, for epistemic uncertainty, dropout was set to 0.5 across the \texttt{VGG} model and for aleatoric uncertainty the same transformations as during training are employed.

\subsection{Results}\label{appendix:results}

\begin{figure}[ht]
    \begin{subfigure}{0.25\textwidth}
        \includegraphics[width=\textwidth]{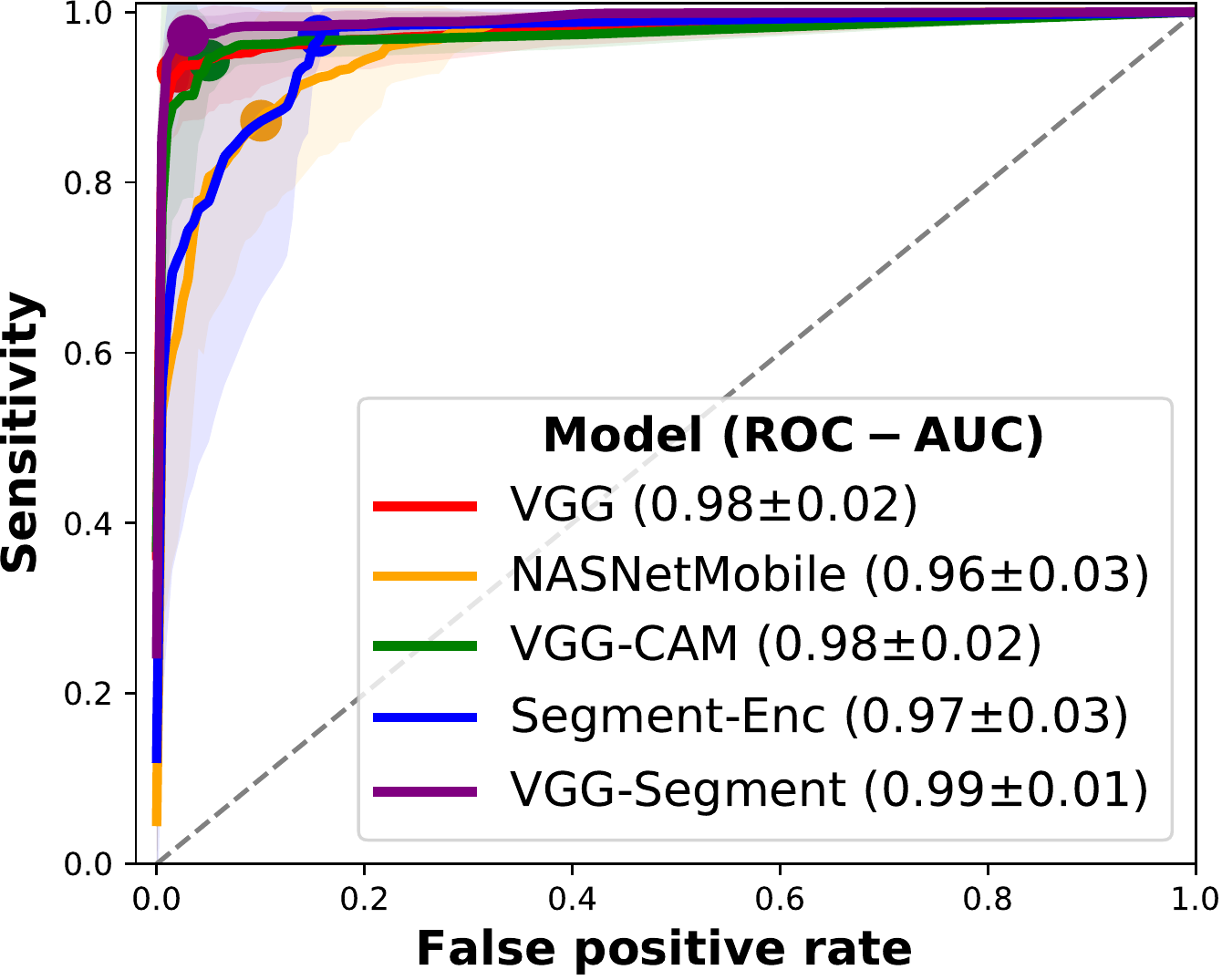}
        \caption{ROC-curve (Pneumonia)}
    \end{subfigure}
    \hfill
    \begin{subfigure}{0.24\textwidth}
        \includegraphics[width=\textwidth]{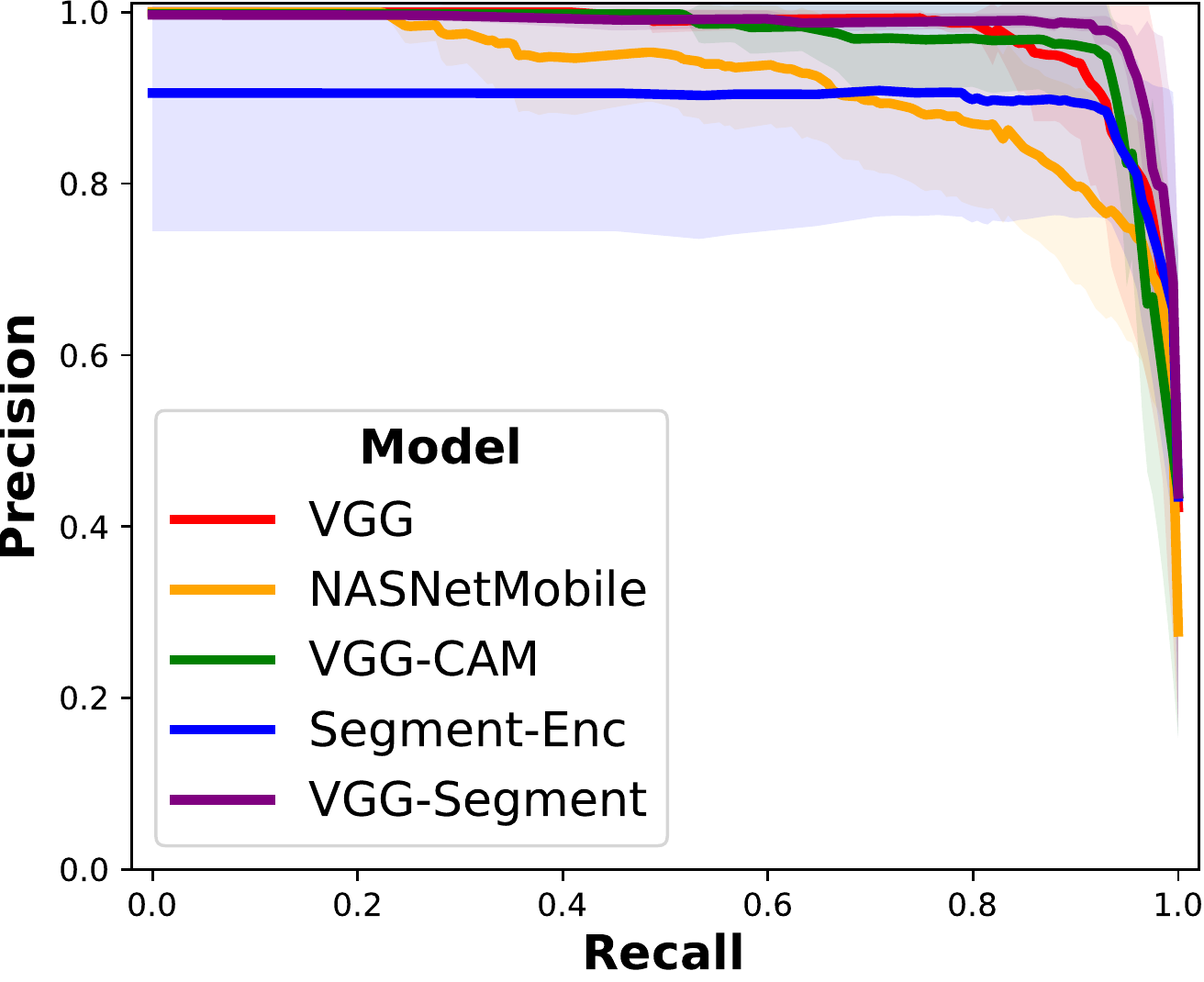}
        \caption{Precision-recall (Pneumonia)}
    \end{subfigure}
    \hfill
    \begin{subfigure}{0.25\textwidth}
        \includegraphics[width=\textwidth]{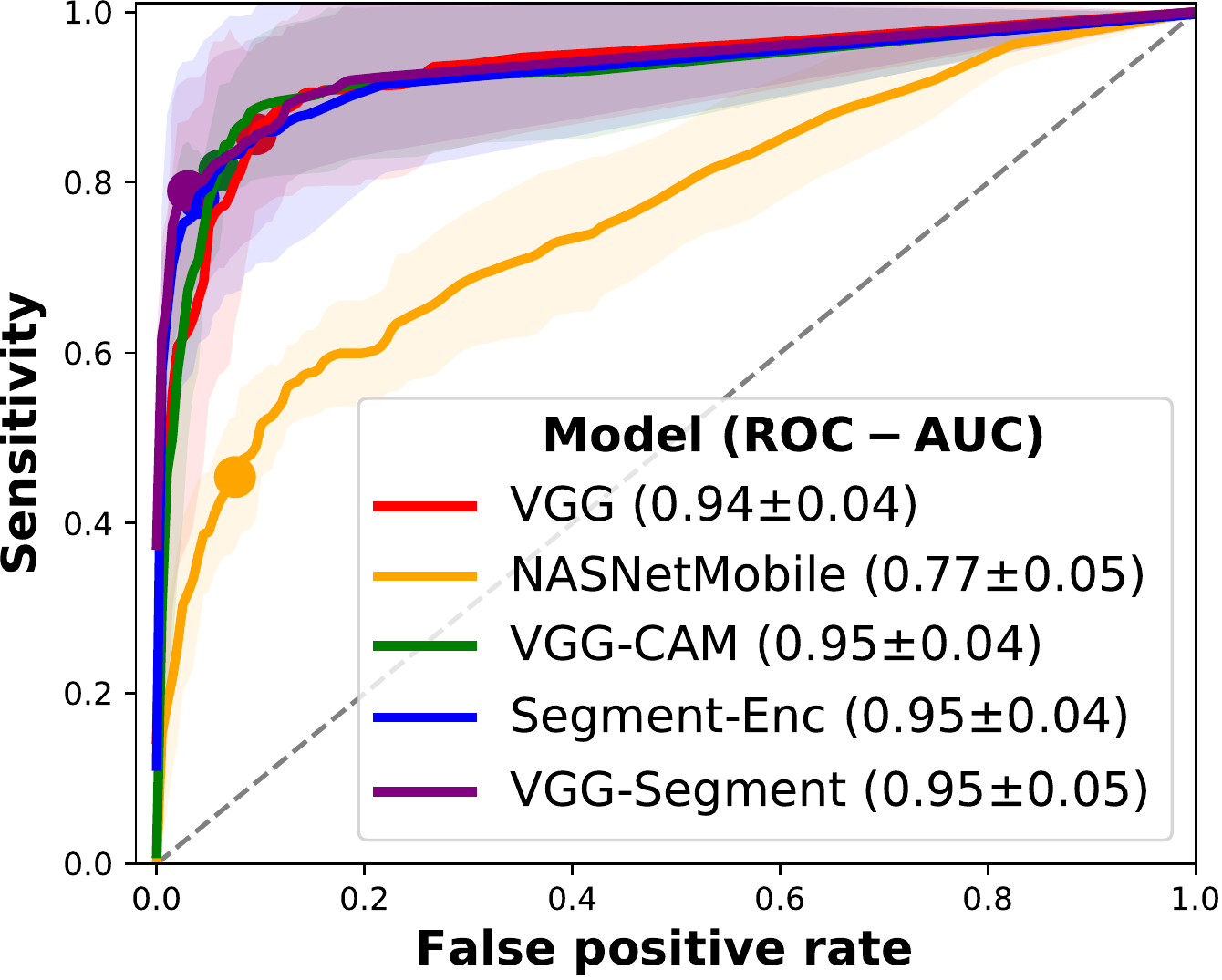}
        \caption{ROC-curve (Healthy)}
    \end{subfigure}
    \hfill
    \begin{subfigure}{0.24\textwidth}
        \includegraphics[width=\textwidth]{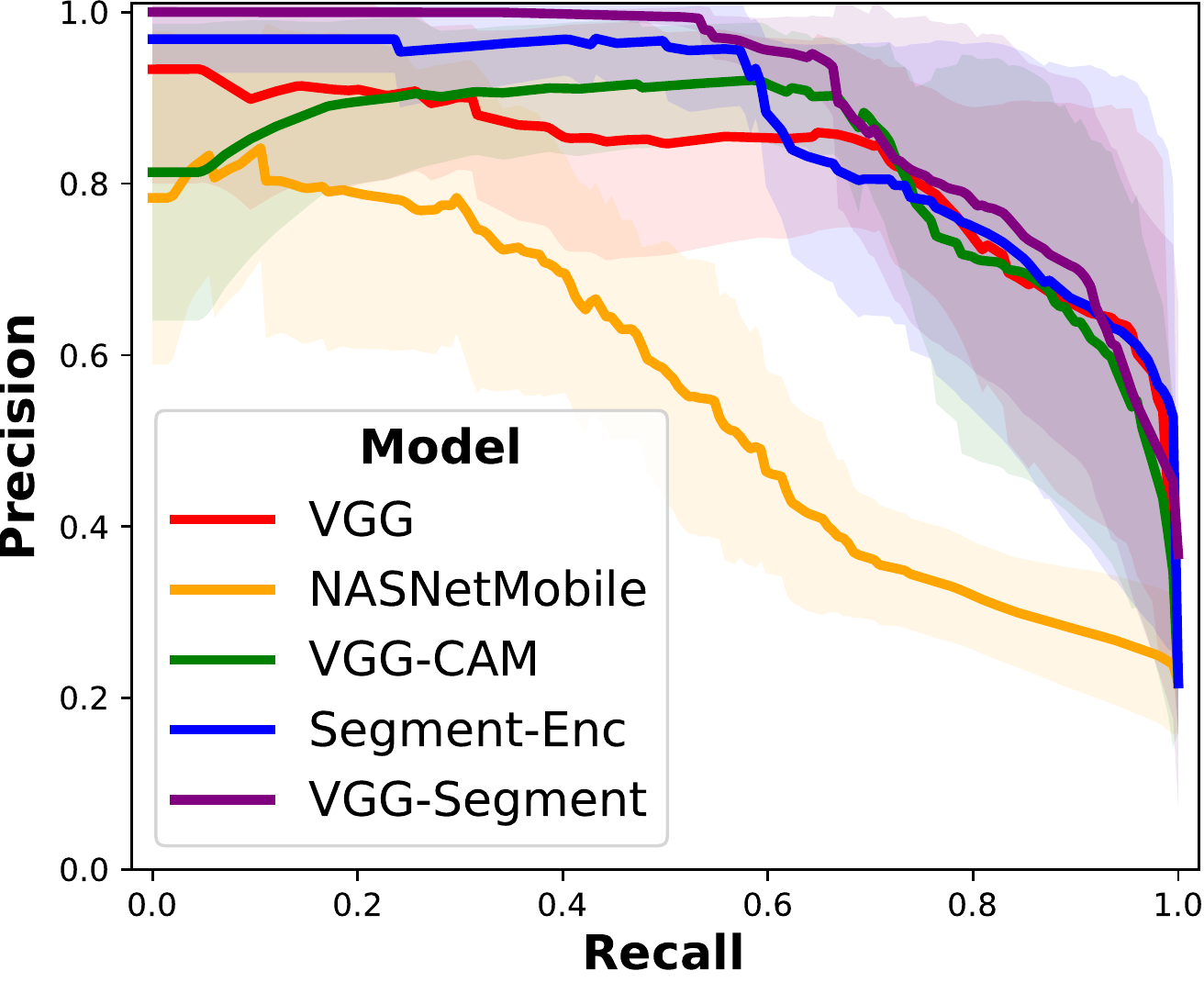}
        \caption{Precision -recall (Healthy)}
    \end{subfigure}
    \caption{\textbf{Binary classification results.} All models achieve good precision and recall in pneumonia detection, but lower scores and higher variances are observed for data of healthy patients} 
    \label{fig:furtherrocs}
\end{figure}

\begin{figure}[ht]
    \begin{subfigure}{0.3\textwidth}
        \includegraphics[width=\textwidth]{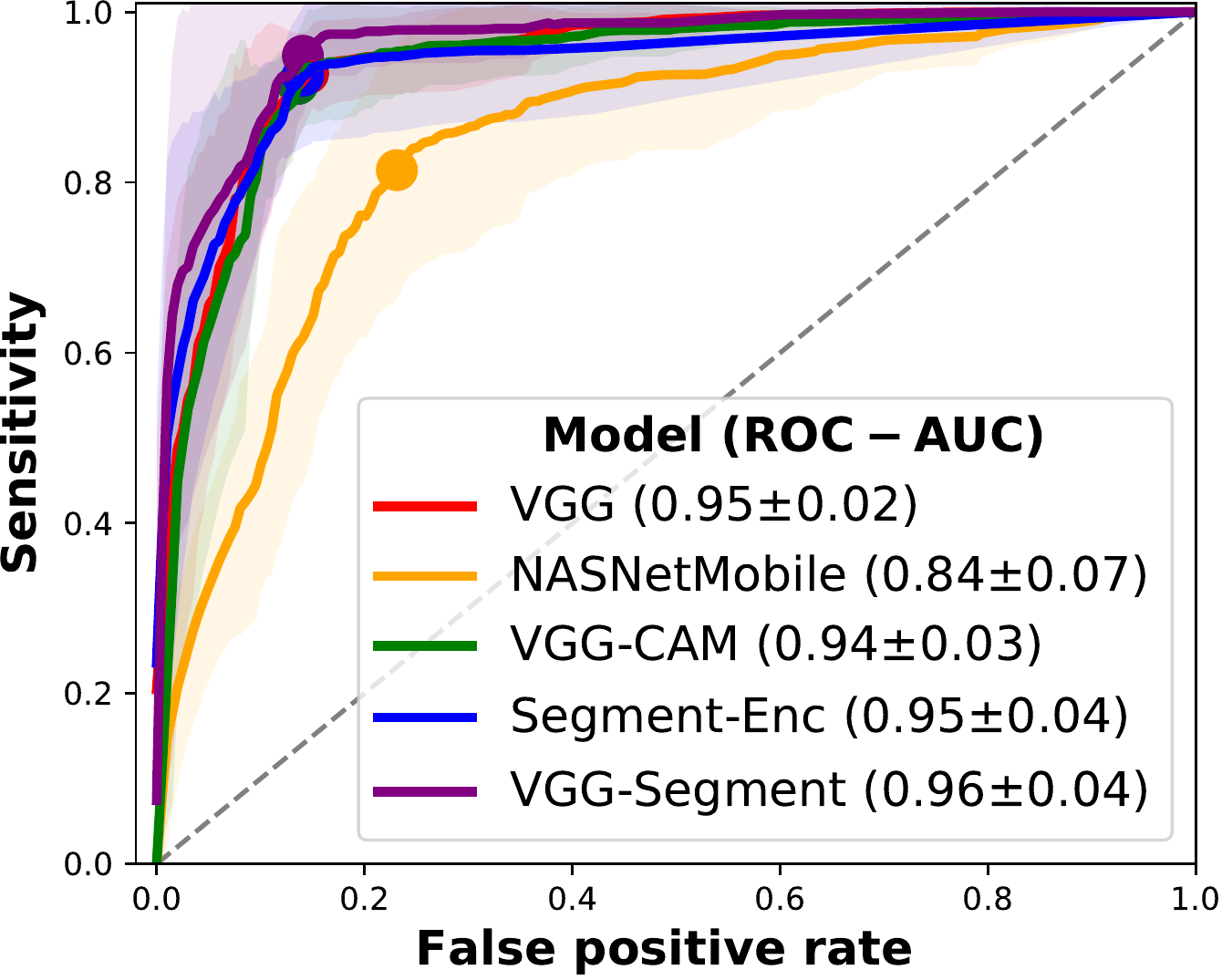}
        \caption{ROC-curve (COVID-19)}
        \label{fig:roccovid}
    \end{subfigure}
    \hfill
    \begin{subfigure}{0.3\textwidth}
        \includegraphics[width=\textwidth]{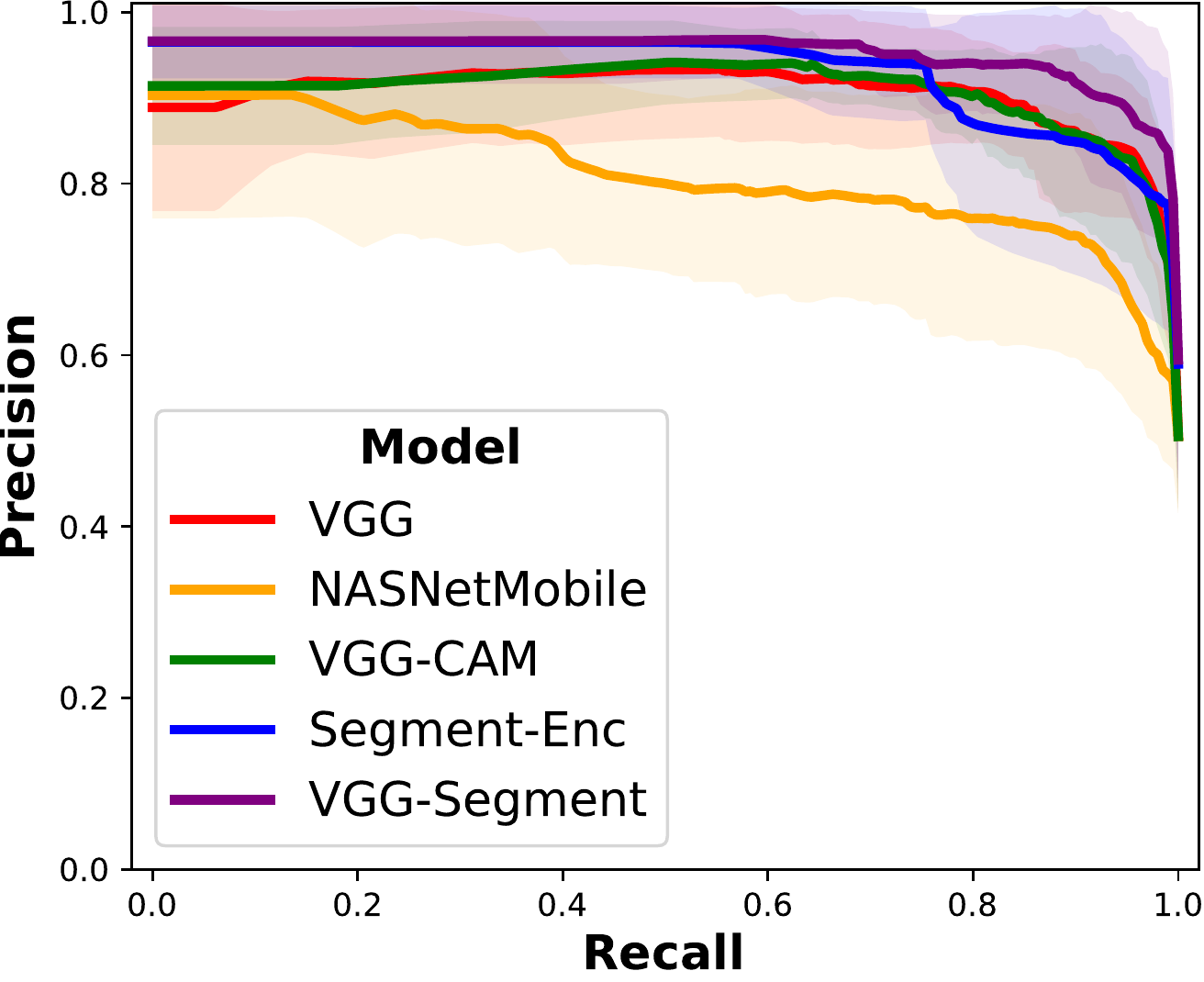}
        \caption{Precision-recall-curve (COVID-19)}
        \label{fig:precreccovid}
    \end{subfigure}
    \hfill
    \begin{subfigure}{0.3\textwidth}
        \includegraphics[width=\textwidth]{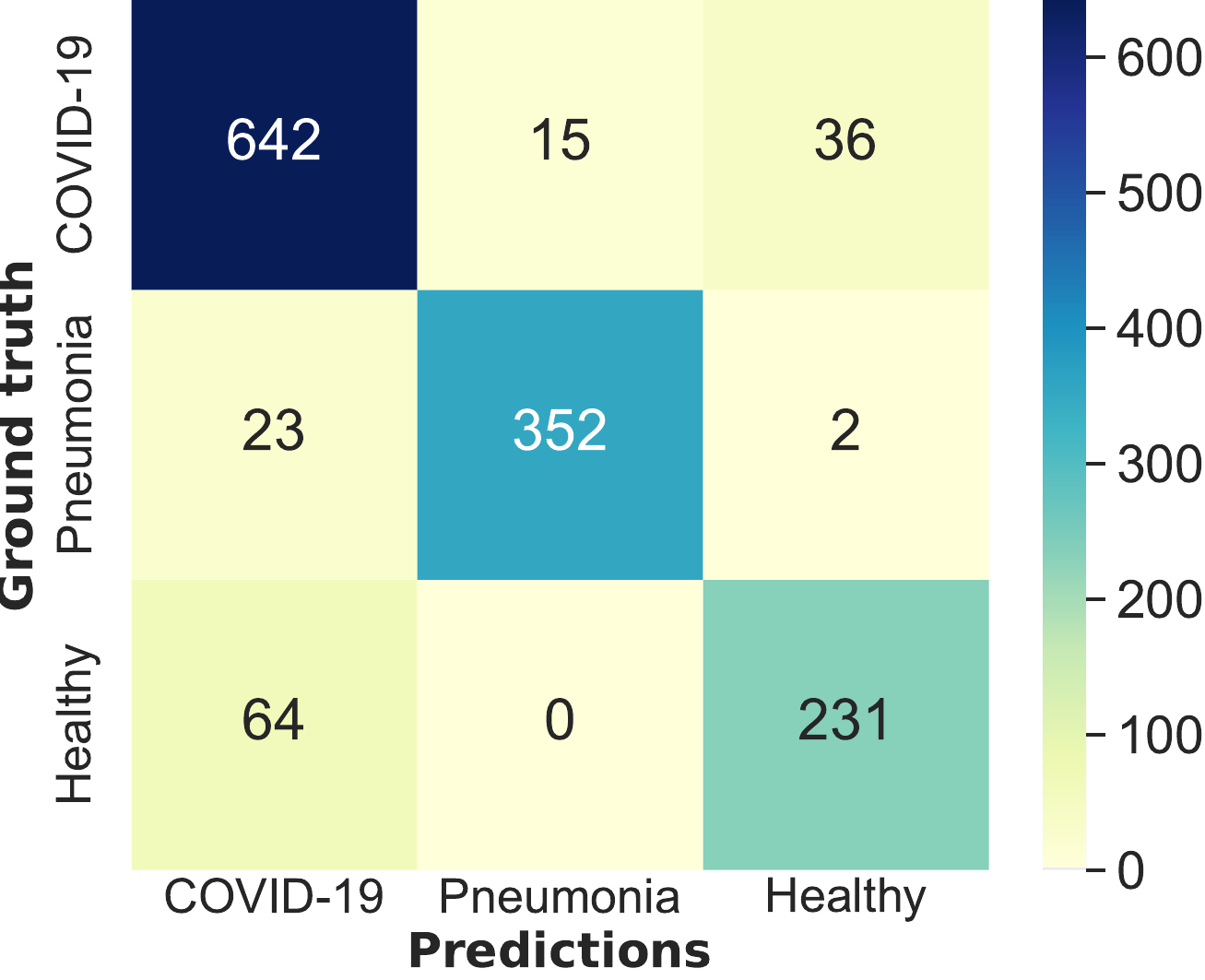}
        \caption{Absolute confusion matrix (\texttt{VGG-CAM})}
        \label{fig:confusion}
    \end{subfigure}
    \caption{\textbf{COVID-19 detection and absolute confusion matrix.}}
    \label{fig:furtherconfusion}
\end{figure}

Re-formulating the classification as a binary task, the ROC-curve and precision-recall curves can be computed for each class. \autoref{fig:furtherrocs} and \autoref{fig:furtherconfusion} depict the performance per class, comparing all proposed models. While pneumonia is distinguished well by all models, \texttt{NASNet} has difficulties with the correct classification of normal lung images. \autoref{fig:roccovid} and \autoref{fig:precreccovid} show that COVID-19 is predicted better than healthy lung images, but not as distinct as pneumonia infections.

Furthermore, in addition to the normalized confusion matrices we provide the absolute values here in \autoref{fig:confusion} (referring to \texttt{VGG-CAM}). Note that most of our data shows COVID-19 infected lungs, despite the novelty of the disease. Problematically, healthy and COVID-19 patients are confused in 100 images, whereas bacterial pneumonia is predicted rather reliably.

\FloatBarrier
\subsubsection{Uninformative class}\label{appendix:uniformative}

Although the main task is defined as differentiating COVID-19, bacterial pneumonia and healthy, we trained the model actually with a fourth ''uninformative'' class in order to identify out-of-distribution samples. This concerns both entirely different pictures (no ultrasound), as well as ultrasound images not showing the lung. Thus, we added 200 images from Tiny ImageNet (one per class taken from the test set) together with 200 neck ultrasound scans taken from the Kaggle ultrasound nerve segmentation challenge. Note that the latter is data recorded with linear ultrasound probes, leading to very different ultrasound images.

\autoref{tab:resultsuninformative} lists the results including these uninformative samples, where better accuracy is achieved due to the ease of distinguishing the uninformative samples from other data. In all cases, precision and recall are higher than 0.98 with low standard deviation.

% \input{uninformative_table}
% Table spreading both columns
\begin{table*}[h]
  \centering
 \scalebox{0.89}{
\begin{tabular}{llccccc}
\toprule
{} &    \textbf{Class} & \textbf{Recall} &  \textbf{Precision} &   \textbf{F1-score} &     \textbf{Specificity} &   \textbf{MCC}  \\
\midrule \Tstrut \Bstrut
\multirow{3}{*}{\pbox{20cm}{\textbf{VGG}\\  \small{Acc.: 0.92, bal.: 0.91} \\ Par.: 14 747 971}
}
& COVID-19 & $ 0.89 \pm {\scriptstyle 0.06 }$ & $ 0.91 \pm {\scriptstyle 0.05 }$ & $ 0.9 \pm {\scriptstyle 0.03 }$ & $ 0.95 \pm {\scriptstyle 0.03 }$ & $ 0.84 \pm {\scriptstyle 0.04 } $ \\
& Pneumonia & $ 0.94 \pm {\scriptstyle 0.05 }$ & $ 0.92 \pm {\scriptstyle 0.06 }$ & $ 0.93 \pm {\scriptstyle 0.05 }$ & $ 0.98 \pm {\scriptstyle 0.02 }$ & $ 0.91 \pm {\scriptstyle 0.06 } $ \\
& Healthy & $ 0.85 \pm {\scriptstyle 0.11 }$ & $ 0.83 \pm {\scriptstyle 0.09 }$ & $ 0.83 \pm {\scriptstyle 0.07 }$ & $ 0.96 \pm {\scriptstyle 0.02 }$ & $ 0.81 \pm {\scriptstyle 0.07 } $ \\
& Uninformative & $ 0.99 \pm {\scriptstyle 0.01 }$ & $ 1.0 \pm {\scriptstyle 0.01 }$ & $ 0.99 \pm {\scriptstyle 0.01 }$ & $ 1.0 \pm {\scriptstyle 0.0 }$ & $ 0.99 \pm {\scriptstyle 0.01 } $ \\
\\\hline \Tstrut \Bstrut
\multirow{3}{*}{\pbox{20cm}{\textbf{VGG-CAM}\\  \small{Acc.: 0.9, bal.: 0.88} \\ Par.: 14 716 227}
} & COVID-19 & $ 0.93 \pm {\scriptstyle 0.05 }$ & $ 0.87 \pm {\scriptstyle 0.07 }$ & $ 0.9 \pm {\scriptstyle 0.05 }$ & $ 0.92 \pm {\scriptstyle 0.04 }$ & $ 0.83 \pm {\scriptstyle 0.07 } $ \\
& Pneumonia & $ 0.93 \pm {\scriptstyle 0.05 }$ & $ 0.95 \pm {\scriptstyle 0.06 }$ & $ 0.94 \pm {\scriptstyle 0.05 }$ & $ 0.99 \pm {\scriptstyle 0.01 }$ & $ 0.92 \pm {\scriptstyle 0.06 } $ \\
& Healthy & $ 0.78 \pm {\scriptstyle 0.1 }$ & $ 0.86 \pm {\scriptstyle 0.08 }$ & $ 0.81 \pm {\scriptstyle 0.05 }$ & $ 0.97 \pm {\scriptstyle 0.02 }$ & $ 0.78 \pm {\scriptstyle 0.05 } $ \\
& Uninformative & $ 0.99 \pm {\scriptstyle 0.01 }$ & $ 0.99 \pm {\scriptstyle 0.01 }$ & $ 0.99 \pm {\scriptstyle 0.01 }$ & $ 1.0 \pm {\scriptstyle 0.0 }$ & $ 0.99 \pm {\scriptstyle 0.01 } $ \\
\\\hline \Tstrut \Bstrut
\multirow{3}{*}{\pbox{20cm}{\textbf{NASNetMobile}\\  \small{Acc.: 0.81, bal: 0.78} \\ Par.: 4 814 487}
}
& COVID-19 & $ 0.87 \pm {\scriptstyle 0.1 }$ & $ 0.74 \pm {\scriptstyle 0.11 }$ & $ 0.8 \pm {\scriptstyle 0.1 }$ & $ 0.82 \pm {\scriptstyle 0.05 }$ & $ 0.67 \pm {\scriptstyle 0.13 } $ \\
& Pneumonia & $ 0.79 \pm {\scriptstyle 0.15 }$ & $ 0.88 \pm {\scriptstyle 0.08 }$ & $ 0.83 \pm {\scriptstyle 0.1 }$ & $ 0.97 \pm {\scriptstyle 0.02 }$ & $ 0.79 \pm {\scriptstyle 0.13 } $ \\
& Healthy & $ 0.47 \pm {\scriptstyle 0.03 }$ & $ 0.61 \pm {\scriptstyle 0.13 }$ & $ 0.53 \pm {\scriptstyle 0.05 }$ & $ 0.94 \pm {\scriptstyle 0.03 }$ & $ 0.46 \pm {\scriptstyle 0.07 } $ \\
& Uninformative & $ 0.99 \pm {\scriptstyle 0.01 }$ & $ 1.0 \pm {\scriptstyle 0.0 }$ & $ 0.99 \pm {\scriptstyle 0.01 }$ & $ 1.0 \pm {\scriptstyle 0.0 }$ & $ 0.99 \pm {\scriptstyle 0.01 } $ \\
\\\hline \Tstrut \Bstrut
\multirow{3}{*}{\pbox{20cm}{\textbf{VGG-Segment}\\  \small{Acc.: 0.93, bal: 0.91} \\ Par.: 34 018 074}
}
& COVID-19 & $ 0.96 \pm {\scriptstyle 0.05 }$ & $ 0.89 \pm {\scriptstyle 0.06 }$ & $ 0.92 \pm {\scriptstyle 0.04 }$ & $ 0.92 \pm {\scriptstyle 0.04 }$ & $ 0.87 \pm {\scriptstyle 0.06 } $ \\
& Pneumonia & $ 0.96 \pm {\scriptstyle 0.03 }$ & $ 0.95 \pm {\scriptstyle 0.03 }$ & $ 0.95 \pm {\scriptstyle 0.02 }$ & $ 0.98 \pm {\scriptstyle 0.01 }$ & $ 0.94 \pm {\scriptstyle 0.03 } $ \\
& Healthy & $ 0.77 \pm {\scriptstyle 0.14 }$ & $ 0.91 \pm {\scriptstyle 0.08 }$ & $ 0.82 \pm {\scriptstyle 0.08 }$ & $ 0.98 \pm {\scriptstyle 0.02 }$ & $ 0.8 \pm {\scriptstyle 0.08 } $ \\
& Uninformative & $ 0.97 \pm {\scriptstyle 0.03 }$ & $ 1.0 \pm {\scriptstyle 0.0 }$ & $ 0.99 \pm {\scriptstyle 0.01 }$ & $ 1.0 \pm {\scriptstyle 0.0 }$ & $ 0.98 \pm {\scriptstyle 0.02 } $ \\
\\\hline \Tstrut \Bstrut
\multirow{3}{*}{\pbox{20cm}{\textbf{Segment-Enc}\\  \small{Acc.: 0.92, bal: 0.91 } \\ Par.: 19 993 307}}
& COVID-19 & $ 0.92 \pm {\scriptstyle 0.09 }$ & $ 0.91 \pm {\scriptstyle 0.06 }$ & $ 0.91 \pm {\scriptstyle 0.03 }$ & $ 0.94 \pm {\scriptstyle 0.04 }$ & $ 0.86 \pm {\scriptstyle 0.04 } $ \\
& Pneumonia & $ 0.95 \pm {\scriptstyle 0.04 }$ & $ 0.89 \pm {\scriptstyle 0.12 }$ & $ 0.92 \pm {\scriptstyle 0.07 }$ & $ 0.96 \pm {\scriptstyle 0.04 }$ & $ 0.9 \pm {\scriptstyle 0.08 } $ \\
& Healthy & $ 0.79 \pm {\scriptstyle 0.17 }$ & $ 0.89 \pm {\scriptstyle 0.1 }$ & $ 0.82 \pm {\scriptstyle 0.11 }$ & $ 0.98 \pm {\scriptstyle 0.01 }$ & $ 0.81 \pm {\scriptstyle 0.12 } $ \\
& Uninformative & $ 1.0 \pm {\scriptstyle 0.0 }$ & $ 1.0 \pm {\scriptstyle 0.0 }$ & $ 1.0 \pm {\scriptstyle 0.0 }$ & $ 1.0 \pm {\scriptstyle 0.0 }$ & $ 1.0 \pm {\scriptstyle 0.0 } $ \\
\bottomrule
\end{tabular}
}
\caption{\textbf{Performance comparison.} Acc. abbreviates accuracy, Bal. balanced accuracy and Par. the number of parameters. The raw results are listed, including the uninformative class. Clearly, this fourth class is very distinctive and is learnt successfully, with almost all scores above 0.89}
\label{tab:resultsuninformative}
\end{table*}

\FloatBarrier
\subsection{Class activation maps}\label{appendix:cams}

\begin{figure}[htb!]
    \centering
        \includegraphics[width=0.8\textwidth]{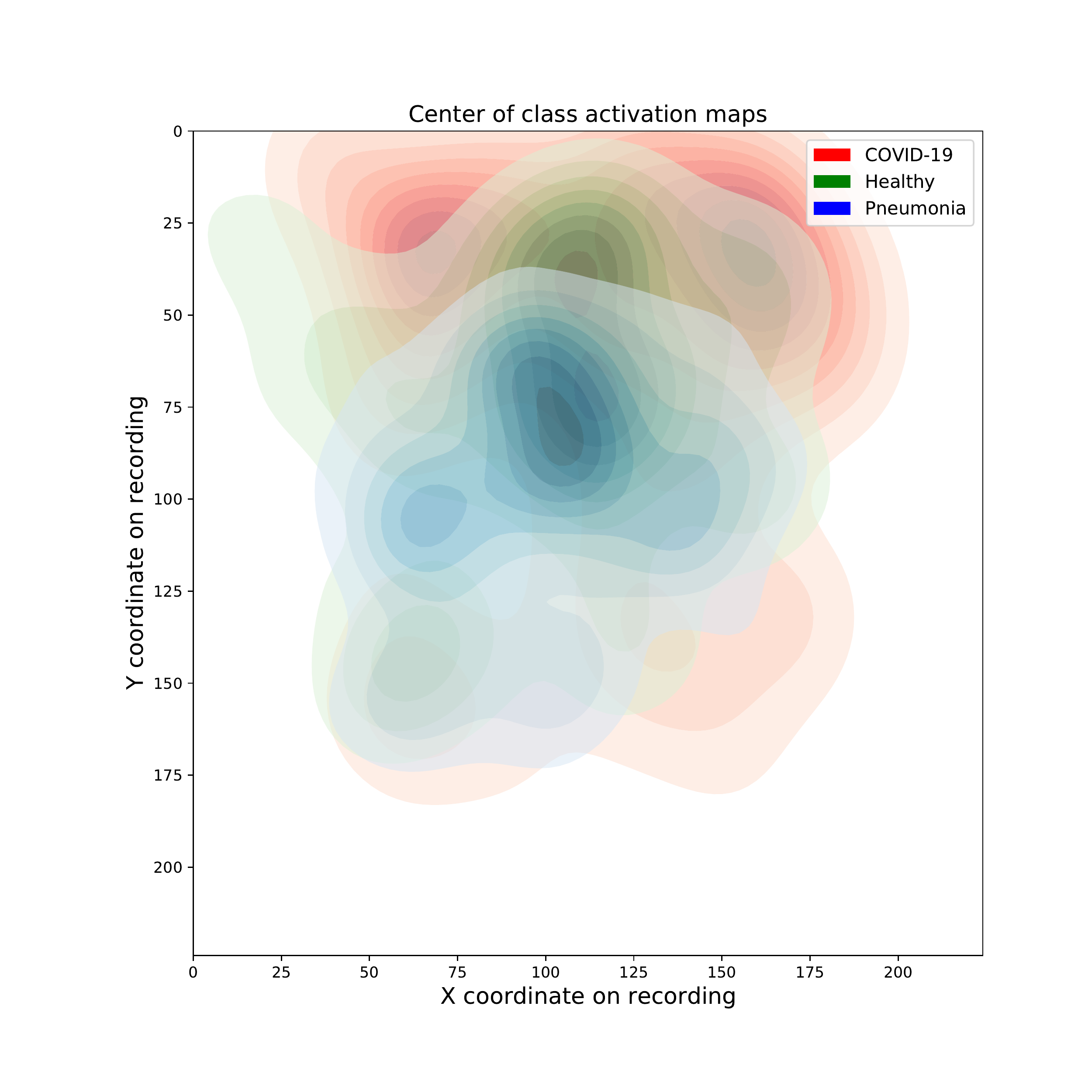}
        \vspace{-4mm}
    \caption{\textbf{Density plot of centers of class activation maps.}
    Pneumonia-CAMs are rather centralized compared to other CAMs. Problematically, COVID-CAMs seem to exhibit a tendency for upper regions of the probe that do not necessarily belong to the lung.}
    \label{fig:camdensity}
\end{figure}

In addition to the scatter plot in~\autoref{fig:cams} we present the corresponding density plot in \autoref{fig:camdensity}, showing the area of the ultrasound image where the class activation is maximal for each class. It can be observed that the activation on healthy and COVID-19 videos is located further in the upper part of the image, where usually only muscles and skin are observed. Further work is thus necessary to analyze and improve the qualitative results of the model.

However, with respect to pathological patterns visible, the model does in many cases focus on the patterns that are interesting to medical experts. \autoref{tab:hmeval} breaks down the results presented in \autoref{fig:patterns} more in detail, and in particular separately for both medical experts. Note that with respect to the pleural line, we only consider the opinion of expert 2 since expert 1 did not mention it. With the exception of consolidations, the difference in responses is quite large, which is however unsurprising for such a qualitative task. Besides the patterns that were already named in \autoref{fig:patterns}, the heatmaps also correctly highlighted air bronchograms (2 cases according to expert 1) and a pleural effusion in 1 out of 7 cases.

% \input{heatmap_table}
% \usepackage{graphicx}
\begin{table}[h]
\centering
\resizebox{\textwidth}{!}{%
\normalsize{
\begin{tabular}{lcccccc}
\hline

& \textbf{Consolidations} & \textbf{A-lines} &  \textbf{B-lines} &   \textbf{Bronchogram} &     \textbf{Effusion} &   \textbf{Pleural line} 
\\ \hline \Tstrut \Bstrut
\textbf{Specific for}                       & \multicolumn{1}{l}{Bacterial pne.}     & \multicolumn{1}{l}{Healthy}          & \multicolumn{1}{l}{COVID-19, viral pne.} & \multicolumn{1}{l}{Bacterial pne.}  & \multicolumn{1}{l}{Pne.}         & \multicolumn{1}{l}{Pne. if irregular} \\ \hline \Tstrut \Bstrut
\textbf{Total visible (expert 1)}    & 18                                           & 13                                    & 12                                             & 2                                         & 7                                      & 20 (expert 2)                                         \\ \hline \Tstrut \Bstrut
\textbf{CAM highlighted (expert 1)} & 17                                           & 6                                     & 0                                              & 2                                         & 1                                      & 0                                           \\ \hline \Tstrut \Bstrut
\textbf{CAM highlighted (expert 2)} & 17                                           & 10                                    & 6                                              & 0                                         & 0                                      & 9                                           \\ \bottomrule \\
\end{tabular}%
}}
\caption{Pathological patterns visible and highlighted by class activation maps of our model. Pneu abbreviated pneumonia. The model focuses on consolidations, A-lines and the pleural line.}
\label{tab:hmeval}
\end{table}
\FloatBarrier
\subsection{Maximum mean discrepancy analysis}
% autoref{fig:MMD bootstrap histograms} shows the results of bootstrapping 

\begin{figure}[h]
    \centering
    \includegraphics[width=0.33\linewidth]{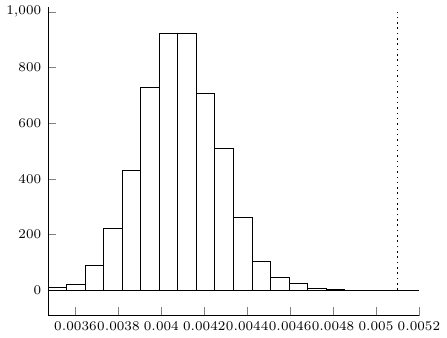}%
    \includegraphics[width=0.33\linewidth]{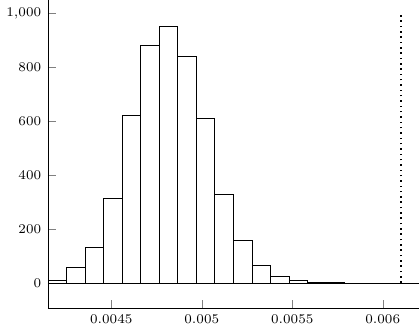}%
    \includegraphics[width=0.33\linewidth]{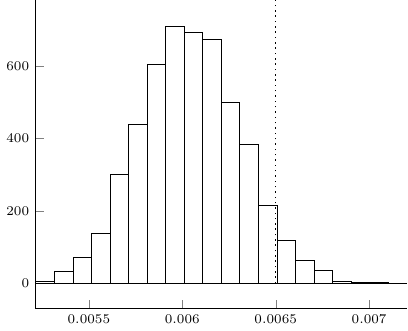}
    \caption{%
        Histograms depicting the empirical null distribution, obtained via bootstrapping $1000$ samples, of the MMD values~(from left to right) $\MMD(\mathbf{C}, \mathbf{P}) \approx 0.0051$,  $\MMD(\mathbf{C}, \mathbf{H}) \approx 0.0061$, 
        and $\MMD(\mathbf{P}, \mathbf{H}) \approx 0.0065$, respectively. The corresponding true MMD values, i.e.\ the ones we obtain by looking at the labels,
        is indicated as a dashed line in each histogram. We observe that these values are highly infrequent under the null distribution, indicating that the differences betwee the three classes are significant.
        Notably, the statistical distance between patients suffering from bacterial pneumonia and healthy patients~(rightmost histogram) achieves a slightly lower empirical significance of $\approx 0.04$. We speculate that this might be related to \emph{other} pre-existing conditions in healthy patients that are not pertinent to this study, though.
    }
    \label{fig:MMD bootstrap histograms}
\end{figure}
\end{appendix}
\end{appendices}
\end{document}